\documentclass[sigconf]{acmart}

\AtBeginDocument{%
  \providecommand\BibTeX{{%
    \normalfont B\kern-0.5em{\scshape i\kern-0.25em b}\kern-0.8em\TeX}}}

\copyrightyear{2023} 
\acmYear{2023} 
\setcopyright{acmlicensed}\acmConference[MM '23]{Proceedings of the 31st ACM International Conference on Multimedia}{October 29-November 3, 2023}{Ottawa, ON, Canada}
\acmBooktitle{Proceedings of the 31st ACM International Conference on Multimedia (MM '23), October 29-November 3, 2023, Ottawa, ON, Canada}
\acmPrice{15.00}
\acmDOI{10.1145/3581783.3612155}
\acmISBN{979-8-4007-0108-5/23/10}

\usepackage{subfig}
\usepackage{bm}
\usepackage{amsthm}
\usepackage{algorithm}
\usepackage{algorithmic}
\usepackage{pifont}
\usepackage{multirow}
\usepackage{array}
\usepackage{booktabs}

\usepackage{bbm}
\usepackage{colortbl}  %彩色表格需要加载的宏包
\usepackage{xcolor}
\usepackage{array} %对表列和表格线的设置需要用到array宏包

\usepackage{subfig}
\usepackage{bm}
\usepackage{amsthm}
\usepackage{algorithm}
\usepackage{algorithmic}
\usepackage{pifont}
\usepackage{multirow}
\usepackage{array}
\usepackage{booktabs}

\usepackage{mathrsfs}
\usepackage{bm}
\usepackage{amsmath}
\usepackage{dsfont}
\usepackage{booktabs}
\usepackage{epstopdf}
\usepackage{multirow}
\usepackage{comment}
\usepackage{endnotes}
\usepackage{enumitem}
\usepackage{makecell}

\renewcommand{\paragraph}[1]{\vspace{1ex}\noindent\textbf{#1}.}

\newcommand{\argmax}{\operatornamewithlimits{argmax}}

\acmSubmissionID{2024}

\begin{document}

\title{Reinforcement Graph Clustering with \\ Unknown Cluster Number}

\author{Yue Liu}
\email{Email: yueliu19990731@163.com}
\affiliation{%
  \institution{NUDT}
  \city{Changsha}
  \state{Hunan}
  \country{China}
}

\author{Ke Liang}
\affiliation{%
  \institution{NUDT}
  \city{Changsha}
  \state{Hunan}
  \country{China}
}

\author{Jun Xia}
\affiliation{%
  \institution{Westlake University}
  \city{Hangzhou}
  \state{Zhejiang}
  \country{China}
}

\author{Xihong Yang}
\affiliation{%
  \institution{NUDT}
  \city{Changsha}
  \state{Hunan}
  \country{China}
}

\author{Sihang Zhou}
\affiliation{%
  \institution{NUDT}
  \city{Changsha}
  \state{Hunan}
  \country{China}
}

\author{Meng Liu}
\affiliation{%
  \institution{NUDT}
  \city{Changsha}
  \state{Hunan}
  \country{China}
}

\author{Xinwang Liu}
\authornote{Corresponding author}
\affiliation{%
  \institution{NUDT}
  \city{Changsha}
  \state{Hunan}
  \country{China}
}

\author{Stan Z. Li}
\affiliation{%
  \institution{Westlake University}
  \city{Hangzhou}
  \state{Zhejiang}
  \country{China}
}

\begin{abstract}
Deep graph clustering, which aims to group nodes into disjoint clusters by neural networks in an unsupervised manner, has attracted great attention in recent years. Although the performance has been largely improved, the excellent performance of the existing methods heavily relies on an accurately predefined cluster number, which is not always available in the real-world scenario. To enable the deep graph clustering algorithms to work without the guidance of the predefined cluster number, we propose a new deep graph clustering method termed Reinforcement Graph Clustering (RGC). In our proposed method, cluster number determination and unsupervised representation learning are unified into a uniform framework by the reinforcement learning mechanism. Concretely, the discriminative node representations are first learned with the contrastive pretext task. Then, to capture the clustering state accurately with both local and global information in the graph, both node and cluster states are considered. Subsequently, at each state, the qualities of different cluster numbers are evaluated by the quality network, and the greedy action is executed to determine the cluster number. In order to conduct feedback actions, the clustering-oriented reward function is proposed to enhance the cohesion of the same clusters and separate the different clusters. Extensive experiments demonstrate the effectiveness and efficiency of our proposed method. The source code of RGC is shared at https://github.com/yueliu1999/RGC and a collection (papers, codes and, datasets) of deep graph clustering is shared at https://github.com/yueliu1999/Awesome-Deep-Graph-Clustering on Github.
\end{abstract}


%% A "teaser" image appears between the author and the affiliation
%% information and the body of the document, and typically spans the
%% page.
%%
%% This command processes the author and affiliation and title
%% information and builds the first part of the formatted document.
\maketitle

\section{Introduction}
In recent years, multimedia techniques have witnessed great success in many domains, like vision \cite{xihong, xiaochang}, language, and graphs. Among various research directions, deep graph clustering is an important unsupervised-learning task to encode nodes in the graph with deep neural networks and separates them into different groups. Motivated by the great success of the graph neural networks \cite{GCN,GAE,GAT} in different areas, such as recommendation \cite{xia2021knowledge}, knowledge graph \cite{AKGR,liang2023message,liang2023abslearn}, and drug prediction \cite{xia2022mole}, anomaly detection \cite{jingcan_1,jingcan_2}, and code search \cite{yingwei_ma}, deep graph clustering methods have been increasingly proposed in recent years. Although achieving promising performance, the recent deep graph clustering algorithms \cite{DAEGC,CCGC,HSAN,SCGC,TGC_ML,arXiv4TGC_ML} are parametric, i.e., requiring the input of predefined cluster number, which is not always available in the real scenario.

% which aims to , 

In the field of traditional clustering \cite{LiangLiTKDE,LSWMKC-2022,zhang2022multiple,wen2023unpaired}, several cluster number estimation methods \cite{review_k} like ELBOW \cite{ELBOW} can be optional to determine the cluster number based on the unsupervised criterion. However, they bring heavy computational costs to deep graph clustering methods since the neural network needs to be repeatedly trained with different cluster numbers and search for the best choice. Experimental evidence can be found in Section \ref{time_efficiency}.

%  1) need manual selections with experience 2) 

%  However, they bring heavy computational cost to deep graph clustering methods since the neural network need to be repeatedly trained with different $K$ and search the best choice. 

\begin{figure}[h]
\centering
\small
\begin{minipage}{1.0\linewidth}
\centerline{\includegraphics[width=1\textwidth]{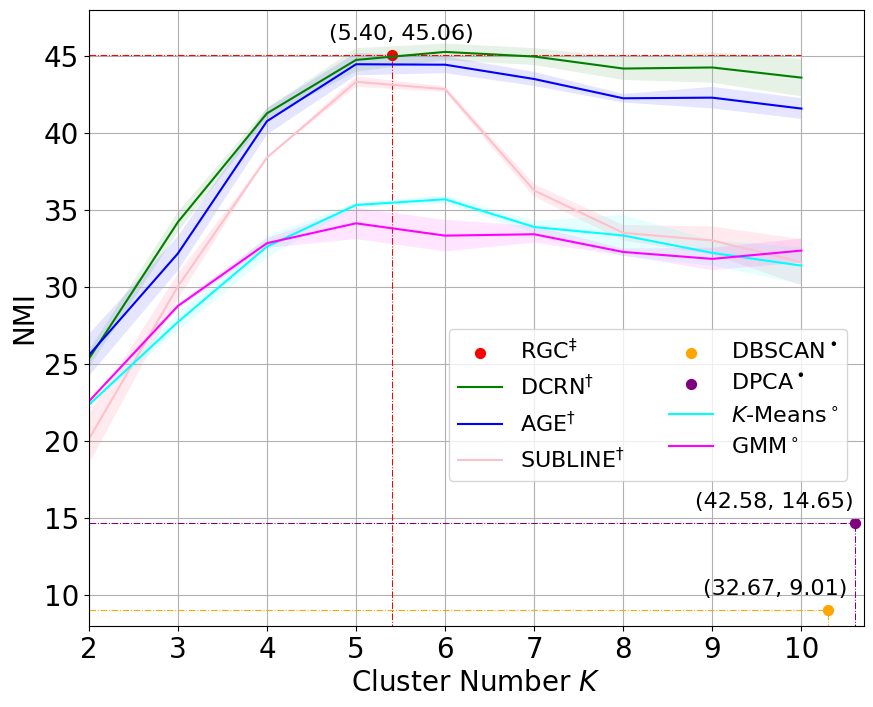}}
\end{minipage}
\caption{Clustering performance comparison on CITESEER dataset. Four categories of the competitors include traditional parametric methods$^\circ$ \cite{K-means,GMM}, traditional non-parametric methods$^\bullet$ \cite{DBSCAN,DPCA}, deep parametric methods$^\dag$ \cite{DCRN,AGE,SUBLIME}, and deep non-parametric method$^\ddag$ (our proposed RGC). For the parametric methods, the results are obtained with the different cluster numbers $K \in [2, 10]$.}
\label{motivation}
\end{figure}

% under different values of cluster number $K$ 

To solve this open problem, a novel deep graph clustering method termed Reinforcement Graph Clustering (RGC) is proposed by learning the cluster number with the reinforcement learning mechanism. Our proposed RGC unifies cluster number determination and unsupervised representation learning into the reinforcement learning framework. Concretely, the nodes are first embedded with the self-supervised encoder, thus improving the discriminative capability of samples. Then, we model the process of determining cluster number as a Markov decision process. To capture the states accurately with both the local and global information in the graph, both node and cluster states are considered. Then, at each state, the designed quality network learns the quality of different cluster numbers, and the greedy action is executed to determine the cluster number. For feedback actions, a clustering-oriented reward function is proposed to enhance the cohesion of the same clusters while separating the different clusters. With the experience replay strategy, the quality network is trained by minimizing the reinforcement learning loss. With these settings, our proposed RGC can automatically determine the cluster number and achieve comparable performance. To demonstrate the effectiveness of our proposed RGC, we conduct experiments as shown in Figure \ref{motivation} and conclude as follows. Firstly, benefiting from the strong unsupervised representation capability of RGC, it significantly outperforms the non-deep parametric/non-parametric methods \cite{K-means,GMM,DBSCAN,DPCA}. Besides, in the reinforcement learning framework, our non-parametric deep method RGC is endowed with the cluster number determination capability, thus achieving comparable performance to the deep parametric methods \cite{DCRN,AGE,SUBLIME}. The main contributions of this work are summarized as follows.

% In summary, our proposed RGC achieves comparable performance even without predefined cluster number $K$. 
% 3) The wrong cluster number $K$ leads to unpromising performance of the deep parametric methods. 

\begin{itemize}[leftmargin=0.5cm]
\item We find the promising performance of recent deep graph clustering methods relies on the predefined cluster number and a new non-parametric deep graph clustering method is proposed.

% We argue that the promising performance of recent deep graph clustering methods relies on the predefined cluster number and propose a new non-parametric deep graph clustering method term RGC.

\item We unify the cluster number determination and the unsupervised learning into a uniform framework by the reinforcement learning mechanism.

\item A clustering-orient reward function is proposed to guide the network to enhance the cohesion of the same clusters and separate the different clusters.

\item Extensive experimental results on five benchmark datasets demonstrate the effectiveness and efficiency of our proposed RGC.

\end{itemize}

\section{RELATED WORK}

\subsection{Deep Graph Clustering}
Attribute graph clustering, which aims to group the nodes in the graph into disjoint clusters, is a fundamental and challenging task. In recent years, benefiting from the strong structural representation capability of graph neural networks \cite{GCN,GAE,GAT,MGCN,10.1145/3606369}, deep graph clustering methods \cite{SDCN,DFCN,DCRN,IDCRN,SCGC} achieve promising performance. According to the learning mechanism, the mainstream methods can be roughly categorized into three classes, i.e., generative methods \cite{MGAE,DAEGC,AGC,GALA,MAGCN,AGCN,SDCN,R-GAE,xiawei_1,liwang_1,liwang_2,MNCI_ML_SIGIR}, adversarial methods \cite{ARGA,ARGA_conf,AGAE,AGC-DRR}, and contrastive methods \cite{MVGRL,AGE,SCAGC,DCRN,IDCRN,GDCL,SCGC,CGC,Dink-Net}. Refer to the survey of deep graph clustering \cite{liuyue_survey}. However, these methods are parametric, i.e., requiring the predefined cluster number $K$. Besides, as shown in Figure \ref{motivation} and Figure \ref{K_analysis}, we observe that the wrong cluster number $K$ will lead to unpromising performance. Although in the field of traditional clustering, several $K$ estimation methods \cite{review_k} like ELBOW \cite{ELBOW} can help researchers to choose the cluster number, they will bring expensive computational costs since the deep neural networks need to be trained for repeated times. The experimental evidence can be found in Figure \ref{time_cost}. To solve the problem, we propose a novel deep graph clustering method termed Reinforcement Graph Clustering (RGC) to automatically determine the cluster number $K$ by reinforcement learning. RGC is the first non-parametric deep graph clustering method. Through extensive experiments, we demonstrate that RGC can save training time compared with the $K$ estimation methods and achieves comparable performance to the state-of-the-art parametric deep graph clustering methods.

% . Experimental evidence can be found in Section \ref{time_efficiency}

% 1) rely on the rich experience of the researchers, 2) 

\subsection{Non-parametric Clustering and Cluster Number Estimation}
In the clustering task, the correct number of clusters is not always known in practice. Thus, there are two alternative options, i.e., cluster number estimation methods or non-parametric clustering methods. The cluster number estimation methods can help to determine the cluster number by performing multiple clustering runs and selecting the best cluster number based on the unsupervised criterion. The mainstream cluster number estimation methods \cite{review_k} include thumb rule, ELBOW \cite{ELBOW}, $t$-SNE \cite{T_SNE}, etc. The thumb rule simply assigns the cluster number $K$ with $\sqrt{N/2}$, where $N$ is the number of samples. This manual setting is empirical and can not be applicable to all datasets. Besides, the ELBOW is a visual method. Concretely, they start the cluster number $K=2$ and keep increasing $K$ in each step by 1, calculating the WSS (within-cluster sum of squares) during training. They choose the value of $K$ when the WSS drops dramatically and after that, it reaches a plateau. However, it will bring large computational costs since the deep neural network needs to be trained with repeated times. Another visual method termed $t$-SNE visualizes the high-dimension data into 2D sample points and helps researchers to determine the cluster number. The effectiveness of $t$-SNE heavily relies on the experience of researchers. Differently, several non-parametric clustering methods, which are free from inputting the cluster number $K$, have been proposed in the field of the traditional clustering \cite{DBSCAN,DPCA} and deep clustering \cite{DEEPDPM}. However, they are not suitable for the graphs since they neglect structural information. Besides, there are few deep non-parametric graph clustering methods. From this motivation, we focus on making deep graph clustering methods work without the guidance of the predefined cluster number. 

% In the field of the traditional clustering, there contains several 

\section{Reinforcement Graph Clustering}
\begin{figure*}
\centering
\includegraphics[scale=0.56]{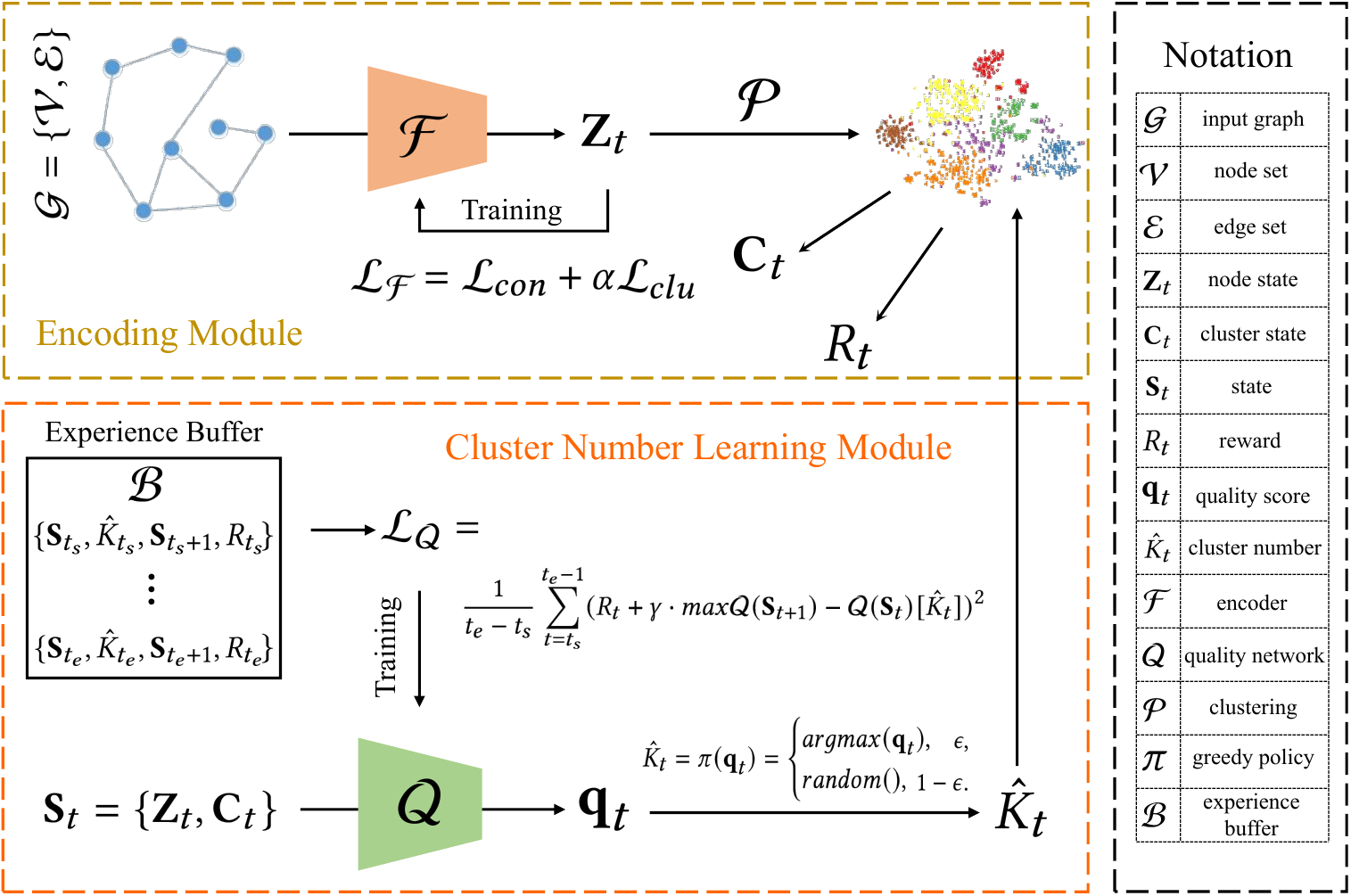}
\caption{Reinforcement Graph Clustering (RGC). In a graph $\mathcal{G}$, the nodes are encoded with $\mathcal{F}$, which is trained in a self-supervised manner, thus improving the discriminative capability of samples. Then, to capture local and global information, state $\textbf{S}_t$ is built with both node and cluster states. In addition, the quality network $\mathcal{Q}$ is designed to evaluate cluster numbers (actions) at $\textbf{S}_t$ and take the greedy action under policy $\pi$. Moreover, a clustering-oriented reward $R_t$ is proposed to improve the cohesion of the same clusters while separating different clusters. With the experience buffer $\mathcal{B}$, $\mathcal{Q}$ is trained by minimizing $\mathcal{L}_{\mathcal{Q}}$. By these settings, our RGC can intelligently determine the cluster number and achieve promising clustering performance.}
\label{overall}  
\end{figure*}

In this section, we propose a novel deep graph clustering method termed Reinforcement Graph Clustering (RGC) to 
enable the deep graph clustering algorithm to work without the guidance of the predefined cluster number. In RGC, a cluster number learning module is designed based on reinforcement learning to automatically determine the cluster number. Besides, the discriminative node representations are learned in a self-supervised manner. By these settings, RGC can intelligently determine the cluster number and achieve promising clustering performance. The overall framework is demonstrated in Figure \ref{overall}. In the following sections, we will introduce the notations, define the problem, and detail the proposed RGC.

% To make deep graph clustering method work without predefined cluster number, RGC is proposed to learn cluster number $\hat{K}$ by reinforcement learning. 

% By minimizing $\mathcal{L}_{\mathcal{Q}}$, $\mathcal{Q}$ is guided to intelligently determine the cluster number. 

\subsection{Notation and Problem}
In this paper, $\mathcal{V}=\{v_1, v_2, \dots, v_N\}$ denotes as the node set of $N$ nodes with $K$ classes. Besides, $\mathcal{E}$ denotes a set of edges in the graph $\mathcal{G} = \{\mathcal{V}, \mathcal{E}\}$. In the matrix form, $\textbf{X} \in \mathds{R}^{N\times D}$ and $\textbf{A} \in \mathds{R}^{N\times N}$ denotes the attribute matrix and the original adjacency matrix, respectively. The notations are summarized in Table \ref{NOTATION_TABLE}. The target of deep graph clustering is to encode nodes with neural networks $\mathcal{F}$ in an unsupervised manner and then divide them into several disjoint clusters. In the existing deep graph clustering methods, the promising performance heavily relies on the precisely predefined cluster number $K$, which is not always known in the real scenario. Thus, this work focuses on making the deep graph clustering method work without the guidance of the predefined cluster number $K$.

\begin{table}[!t]
\centering
\small
\scalebox{1.3}{
\begin{tabular}{ll}
\toprule
\textbf{Notation}                                        & \textbf{Meaning}                                \\ \midrule
$\textbf{X}\in \mathds{R}^{N\times D}$  & Attribute matrix  
\\
$\textbf{A}\in \mathds{R}^{N\times N}$  & Original adjacency matrix   
\\
$\textbf{L}\in \mathds{R}^{N\times N}$  & Graph Laplacian matrix
\\
$\textbf{Z} \in \mathds{R}^{N\times d}$   & Node embeddings
\\
$\textbf{C} \in \mathds{R}^{\hat{K}\times d}$   & Cluster embeddings
\\
$\textbf{Z}_t \in \mathds{R}^{N\times d}$   & Node state at epoch $t$      
\\
$\textbf{C}_t \in \mathds{R}^{\hat{K}\times d}$   & Cluster state at epoch $t$      
\\
$\hat{K}_t \in \mathds{R}$   & Learned cluster number at epoch $t$
\\
$\textbf{S}_t=\{\textbf{Z}_t,\textbf{C}_t\}$   & State at epoch $t$
\\
$\textbf{q}_t \in \mathds{R}^{N_K}$   & Quality score at epoch $t$
\\
$R_t \in \mathds{R}$ & Reward at epoch $t$
\\
$\mathcal{F}$   & Sample Encoder Network
\\
$\mathcal{Q}$   & Quality Network
\\
$\pi$   & Policy with $\epsilon$-greedy
\\
\bottomrule
\end{tabular}
}
\caption{Basic notation summary.}
\label{NOTATION_TABLE} 
\end{table}

% \\
% $\widetilde{\textbf{L}}\in \mathds{R}^{N\times N}$  & Symmetric normalized Laplacian matrix
% Following previous contrastive deep graph clustering methods, 

\subsection{Encoding}
Firstly, we encode the nodes to the embeddings $\textbf{Z} \in \mathds{R}^{N \times d}$ with the encoder $\mathcal{F}$ as follow:

\begin{equation} 
\textbf{Z} = \mathcal{F}(\textbf{X}, \textbf{A}),
\label{encoding}
\end{equation}
where $\textbf{X}$ and $\textbf{A}$ denote the attribute matrix and adjacency matrix, respectively. The detailed designation of $\mathcal{F}$ is described in Appendix A.1. The encoder is trained self-supervised. Concretely, the encoder loss $\mathcal{L}_{\mathcal{F}}$ contains the contrastive loss and the clustering loss described in Section \ref{training}. After encoding, the clustering algorithm $\mathcal{P}$ will be performed on $\textbf{Z}$ and obtain the clustering results as follow:

\begin{equation} 
\textbf{C}, \textbf{P} = \mathcal{P}(\textbf{Z}, \hat{K}),
\label{encoding}
\end{equation}
where $\hat{K}$ denotes the learned cluster number and $\textbf{P} \in \mathds{R}^{N \times \hat{K}}$ denotes the cluster assignment matrix. Besides, the cluster embeddings $\textbf{C} \in \mathds{R}^{\hat{K} \times d}$ are calculated by averaging the node embeddings within each cluster. In the next section, we detail the process of learning cluster number $\hat{K}$

\subsection{Cluster Number Learning Module}
In this section, the cluster number learning module is proposed based on reinforcement learning. Firstly, the cluster number determination is modelled as the Markov decision process during training. Four important elements in the Markov decision process, including state, action, transition, and reward, are defined as follows.

\begin{itemize}[leftmargin=0.5cm]
\item State. 
In order to capture the local and global information in the graph, the state $\textbf{S}_t$ at $t$ epoch are built with both the node representations and the cluster representations. Formally, the states are formulated as follows:
\begin{equation} 
\textbf{S}_t = \{\textbf{Z}_t,\textbf{C}_t\},
\label{encoding}
\end{equation}
where $\textbf{C}_t \in \mathds{R}^{\hat{K} \times d}$ is calculated by averaging the node embeddings within each cluster. In this manner, the states will contain both the node and cluster information, better revealing the potential semantics in the graph.

\item Transition.
The state $\textbf{S}_t = \{\textbf{Z}_t,\textbf{C}_t\}$ will be transitioned to $\textbf{S}_{t+1} = \{\textbf{Z}_{t+1},\textbf{C}_{t+1}\}$ in the process of training. Concretely, when the training epoch number increases, the encoder $\mathcal{F}$ will be optimized by minimizing encoder loss $\mathcal{L}_{\mathcal{F}}$, leading to a state transition. 

\item Action.
To evaluate the quality of different cluster numbers, the quality network is carefully designed as follows: 
\begin{equation} 
\textbf{q}_t = \mathcal{Q}(\textbf{S}_t) = \mathcal{Q}(\{\textbf{Z}_t, \textbf{C}_t\}),
\label{quality}
\end{equation}
where $\textbf{q}_t \in \mathds{R}^{N_K}$ denotes the quality score vector and $\textbf{q}_t[i]$ indicates the quality of cluster number $i$. Besides, $N_k$ is the max cluster number. Based on $\textbf{q}_t$, the cluster number is estimated as follows:
\begin{equation} 
\hat{K}_t = \pi(\textbf{q}_t) =  
\begin{cases}
\argmax(\textbf{q}_t), & \epsilon, \\
\text{random}(), & 1-\epsilon.
\end{cases}
\label{action}
\end{equation}
Here, the $\epsilon$-greedy strategy is adopted. To be precise, under the policy $\pi$, $\mathcal{Q}$ takes the greedy action with the probability $\epsilon$ and random decision with the probability $1-\epsilon$. The greedy parameter $\epsilon$ increases as the training epoch number increases, guiding the policy $\pi$ to take intelligent decision with higher probability. In this way, our proposed method RGC can evaluate different cluster numbers and takes the intelligent decision during training. The detailed designation of the quality network can be found in Appendix A.3.

% the states with 

% The action is to select the estimated cluster number $\hat{K}$. 

% Besides, the $\mathcal{Q}$.

% $K_{max}$ is the max choice number of the cluster number. 

\item Reward. After taking an action at state $\textbf{S}_t$, the reward function will feedback to the network. In our proposed RGC, for the clustering task, a clustering-oriented reward function $\mathcal{R}$ is proposed as follows. 
\begin{equation} 
\begin{aligned}
R_t = \mathcal{R}(\textbf{S}_t,\hat{K}_t)= \mathcal{R}(\{\textbf{Z}_t, \textbf{C}_t\},\hat{K}_t) &= \\ - \frac{1}{N} \sum_i \min_j  \mathcal{D}(\textbf{Z}_t[i], \textbf{C}_t[j])  + \frac{1}{\hat{K}^2} \sum_i \sum_j & \mathcal{D}(\textbf{C}_{t}[i], \textbf{C}_{t}[j]),
\end{aligned}
\label{reward}
\end{equation}
where $\mathcal{D}$ denotes the distance metric function like Euclidean distance. During training, our quality network is encouraged to maximize the reward $R_t \in \mathds{R}$. Through $max R_t$, the first term in Eq. \eqref{reward} minimizes the distance between each node and the corresponding cluster center while the second term in Eq. \eqref{reward} maximizes the distance between different cluster centers. Guided by the clustering-oriented reward function, the proposed RGC can improve cohesion in the same clusters while separating different clusters.

\end{itemize}

In these settings, the Markov decision process of determining the cluster number is constructed. Subsequently, the experiences are collected by the experience replay strategy and the quality network is trained by minimizing the reinforcement loss. In the next section, we detail the training of our proposed method. 

\begin{table*}[!t]
\caption{Clustering performance of the state-of-the-art deep parametric methods $^\dag$ and our proposed deep non-parametric method RGC$^\ddag$. All results are obtained by ten runs and reported with mean±std (\%). The bold and underlined result denotes the champion and runner-up.}
\scalebox{0.8}{
\begin{tabular}{cc|cc|cc|cc|cc|cc}
\hline
\multicolumn{2}{c|}{\multirow{2}{*}{\textbf{Method}}} & \multicolumn{2}{c|}{\textbf{BAT}}         & \multicolumn{2}{c|}{\textbf{EAT}}         & \multicolumn{2}{c|}{\textbf{AMAP}}        & \multicolumn{2}{c|}{\textbf{CITESEER}}    & \multicolumn{2}{c}{\textbf{CORA}}         \\ \cline{3-12} 
\multicolumn{2}{c|}{}                                 & \textbf{NMI}        & \textbf{ARI}        & \textbf{NMI}        & \textbf{ARI}        & \textbf{NMI}        & \textbf{ARI}        & \textbf{NMI}        & \textbf{ARI}        & \textbf{NMI}        & \textbf{ARI}        \\ \hline
\textbf{DEC$^\dag$}              & \textbf{ICML'16}          & 14.10±1.99          & 07.99±1.21          & 04.96±1.74          & 03.60±1.87          & 37.35±0.05          & 18.59±0.04          & 28.34±0.30          & 28.12±0.36          & 23.54±0.34          & 15.13±0.42          \\
\textbf{DCN$^\dag$}              & \textbf{ICML'17}          & 18.03±7.73          & 13.75±6.05          & 06.92±2.80          & 05.11±2.65          & 38.76±0.30          & 20.80±0.47          & 27.64±0.08          & 29.31±0.14          & 25.65±0.65          & 21.63±0.58          \\
\textbf{IDEC$^\dag$}             & \textbf{IJCAI'17}         & 12.80+1.74          & 07.85+1.31          & 04.63+0.97          & 03.19+0.76          & 37.83+0.08          & 19.24+0.07          & 27.17+2.40          & 25.70+2.65          & 26.31+1.22          & 22.07+1.53          \\
\textbf{AdaGAE$^\dag$}           & \textbf{TPAMI'21}         & 15.84±0.78          & 07.80±0.41          & 04.36±1.87          & 02.47±0.54          & 55.96±0.87          & 46.20±0.45          & 27.79±0.47          & 24.19±0.85          & 32.19±1.34          & 28.25±0.98          \\ \hline
\textbf{MGAE$^\dag$}             & \textbf{CIKM'17}          & 30.59±2.06          & 24.15±1.70          & 20.69±0.98          & 18.33±1.79          & 62.13±2.79          & 48.82±4.57          & 34.63±0.65          & 33.55±1.18          & 28.78±2.97          & 16.43±1.65          \\
\textbf{DAEGC$^\dag$}            & \textbf{IJCAI'19}         & 21.43±0.35          & 18.18±0.29          & 21.33±0.44          & 20.50±0.51          & 65.25±0.45          & 58.12±0.24          & 36.41±0.86          & 37.78±1.24          & 52.89±0.69          & 49.63±0.43          \\
\textbf{ARGA$^\dag$}             & \textbf{IJCAI'19}         & {\underline{49.09±0.54}}    & {\underline{42.02±1.21}}    & 25.44±0.31          & 16.57±0.31          & 58.36±2.76          & 44.18±4.41          & 34.40±0.71          & 34.32±0.70          & 51.06±0.52          & 47.71±0.33          \\
\textbf{SDCN$^\dag$}             & \textbf{WWW'20}           & 25.74±5.71          & 21.04±4.97          & 21.61±1.26          & 21.63±1.49          & 44.85±0.83          & 31.21±1.23          & 38.71±0.32          & 40.17±0.43          & 14.28±1.91          & 07.78±3.24          \\
\textbf{AGE$^\dag$}              & \textbf{KDD'20}           & 36.04±1.54          & 26.59±1.83          & 23.64±0.66          & 20.39±0.70          & 65.38±0.61          & 55.89±1.34          & 44.93±0.53          & 45.31±0.41          & \underline{57.58±1.42} & {\underline{50.10±2.14}}    \\
\textbf{MVGRL$^\dag$}            & \textbf{ICML'20}          & 29.33±0.70          & 13.45±0.03          & 21.53±0.94          & 17.12±1.46          & 30.28±3.94          & 18.77±2.34          & 40.69±0.93          & 34.18±1.73          & 55.57±1.54          & 48.70±3.94          \\
\textbf{DFCN$^\dag$}             & \textbf{AAAI'21}          & 48.77±0.51          & 37.76±0.23          & {\underline{26.49±0.41}}    & 11.87±0.23          & 66.23±1.21          & 58.28±0.74          & 43.90±0.20          & 45.50±0.30          & 19.36±0.87          & 04.67±2.10          \\
\textbf{GDCL$^\dag$}             & \textbf{IJCAI'21}         & 31.70±0.42          & 19.33±0.57          & 25.10±0.01          & {\underline{21.76±0.01}}    & 37.32±0.28          & 21.57±0.51          & 39.52±0.38          & 41.07±0.96          & 56.60±0.36          & 48.05±0.72          \\
\textbf{GCA$^\dag$}              & \textbf{WWW'21}           & 38.88±0.23          & 26.69±2.85          & 24.05±0.25          & 14.37±0.19          & 48.38±2.38          & 26.85±0.44          & 36.15±0.78          & 35.20±0.96          & 46.87±0.65          & 30.32±0.98          \\
\textbf{MCGC$^\dag$}             & \textbf{NIPS'21}          & 23.11+0.56          & 8.41+0.32           & 16.64+0.41          & 12.21+0.13          & \multicolumn{2}{c|}{OOM}                  & 39.11+0.06          & 37.54+0.12          & 24.11+1.00          & 14.33+1.26          \\
\textbf{AutoSSL$^\dag$}          & \textbf{ICLR'22}          & 17.84±0.98          & 13.11±0.81          & 17.86±0.22          & 13.13±0.71          & 48.56±0.71          & 26.87±0.34          & 40.67±0.84          & 38.73±0.55          & 47.62±0.45          & 38.92±0.77          \\
\textbf{AGC-DRR$^\dag$}          & \textbf{IJCAI'22}         & 19.91±0.24          & 14.59±0.13          & 11.15±0.24          & 09.50±0.25          & {\underline{66.54±1.24}}    & \textbf{60.15±1.56} & 43.28±1.41          & 45.34±2.33          & 18.74±0.73          & 14.80±1.64          \\
\textbf{DCRN$^\dag$}             & \textbf{AAAI'22}          & 47.23±0.74          & 39.76±0.87          & 24.09±0.53          & 17.17±0.69          & \multicolumn{2}{c|}{OOM}                  & \textbf{45.86±0.35} & \textbf{47.64±0.30} & 45.13±1.57          & 33.15±0.14          \\
\textbf{AFGRL$^\dag$}            & \textbf{AAAI'22}          & 27.55±0.62          & 21.89±0.74          & 17.33±0.54          & 13.62±0.57          & 64.05±0.15          & 54.45±0.48          & 15.17±0.47          & 14.32±0.78          & 12.36±1.54          & 14.32±1.87          \\
\textbf{ProGCL$^\dag$}           & \textbf{ICML'22}          & 28.69+0.92          & 21.84+1.34          & 22.04+2.23          & 14.74+1.99          & 39.56+0.39          & 34.18+0.89          & 39.59+0.39          & 36.16+1.11          & 41.02+1.34          & 30.71+2.70          \\
\textbf{SUBLIME$^\dag$}          & \textbf{WWW'22}           & 22.03±0.48          & 14.45±0.87          & 21.85±0.62          & 19.51±0.45          & 06.37±1.89          & 05.36±2.14          & 43.15±0.14          & 44.21±0.54          & 53.88±1.02          & \textbf{50.15±0.14} \\
\textbf{RGC$^\ddag$}              & \textbf{Ours}             & \textbf{51.58+0.83} & \textbf{47.16+1.35} & \textbf{37.77+0.13} & \textbf{30.16+0.15} & \textbf{69.61+0.36} & {\underline{59.58+0.39}}    & {\underline{45.06+0.93}}    & {\underline{46.17+1.46}}    & {\textbf{57.60+1.36}}    & 49.46+2.72          \\ \hline
\end{tabular}}
\label{compare}
\end{table*}

\begin{figure*}[!t]
\footnotesize
\begin{minipage}{0.139\linewidth}
\centerline{\includegraphics[width=\textwidth]{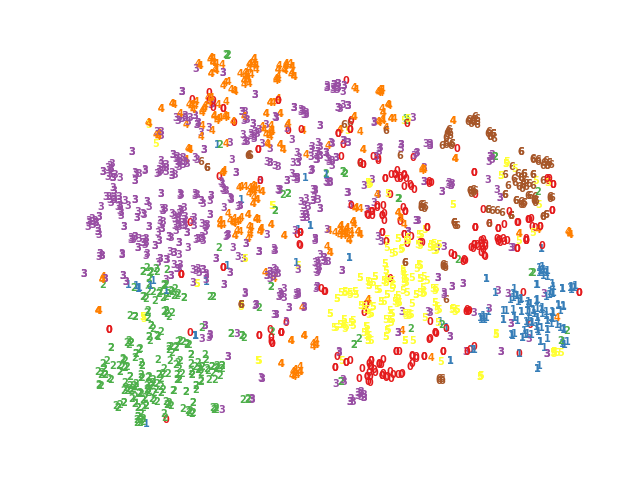}}
\vspace{3pt}
\centerline{\includegraphics[width=\textwidth]{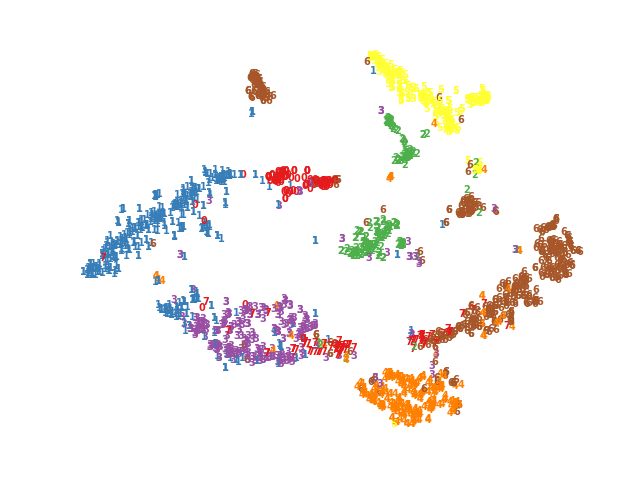}}
\vspace{3pt}
\centerline{(a) DAEGC}
\end{minipage}
\begin{minipage}{0.139\linewidth}
\centerline{\includegraphics[width=\textwidth]{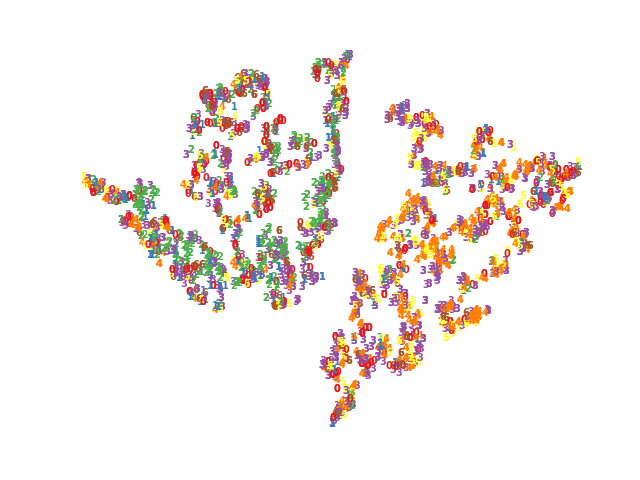}}
\vspace{3pt}
\centerline{\includegraphics[width=\textwidth]{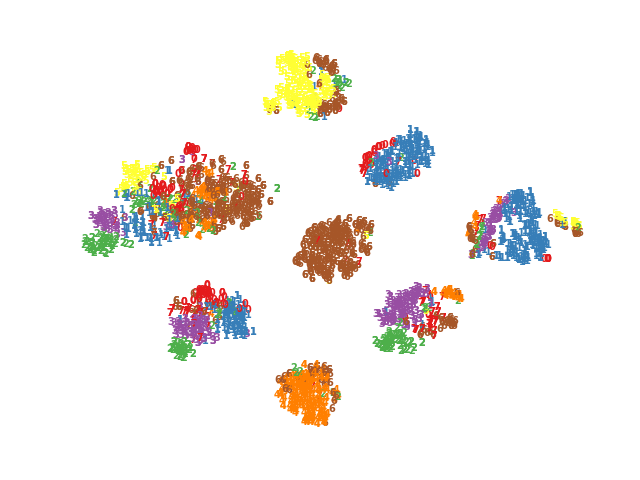}}
\vspace{3pt}
\centerline{(b) SDCN}
\end{minipage}
\begin{minipage}{0.139\linewidth}
\centerline{\includegraphics[width=\textwidth]{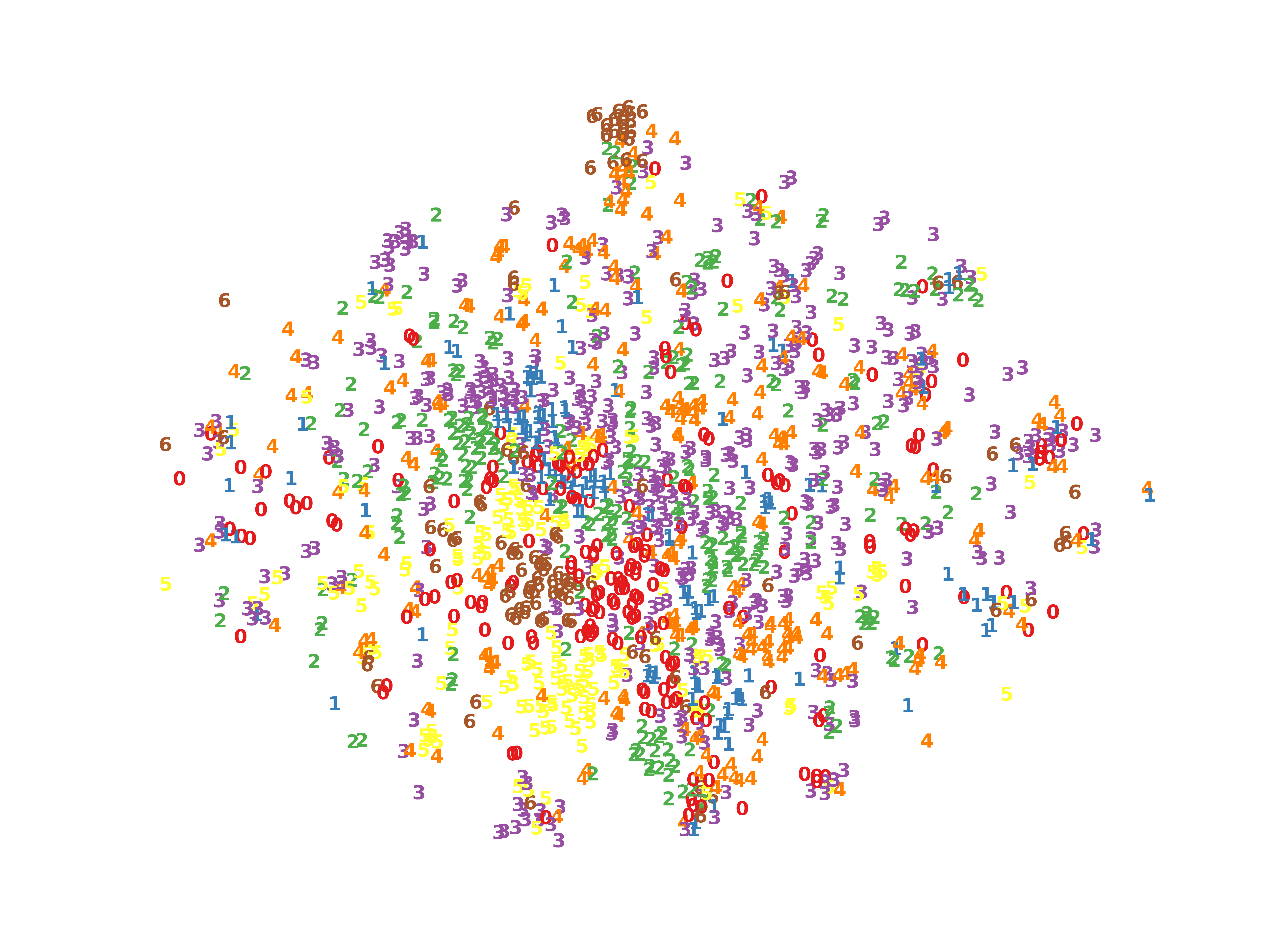}}
\vspace{3pt}
\centerline{\includegraphics[width=\textwidth]{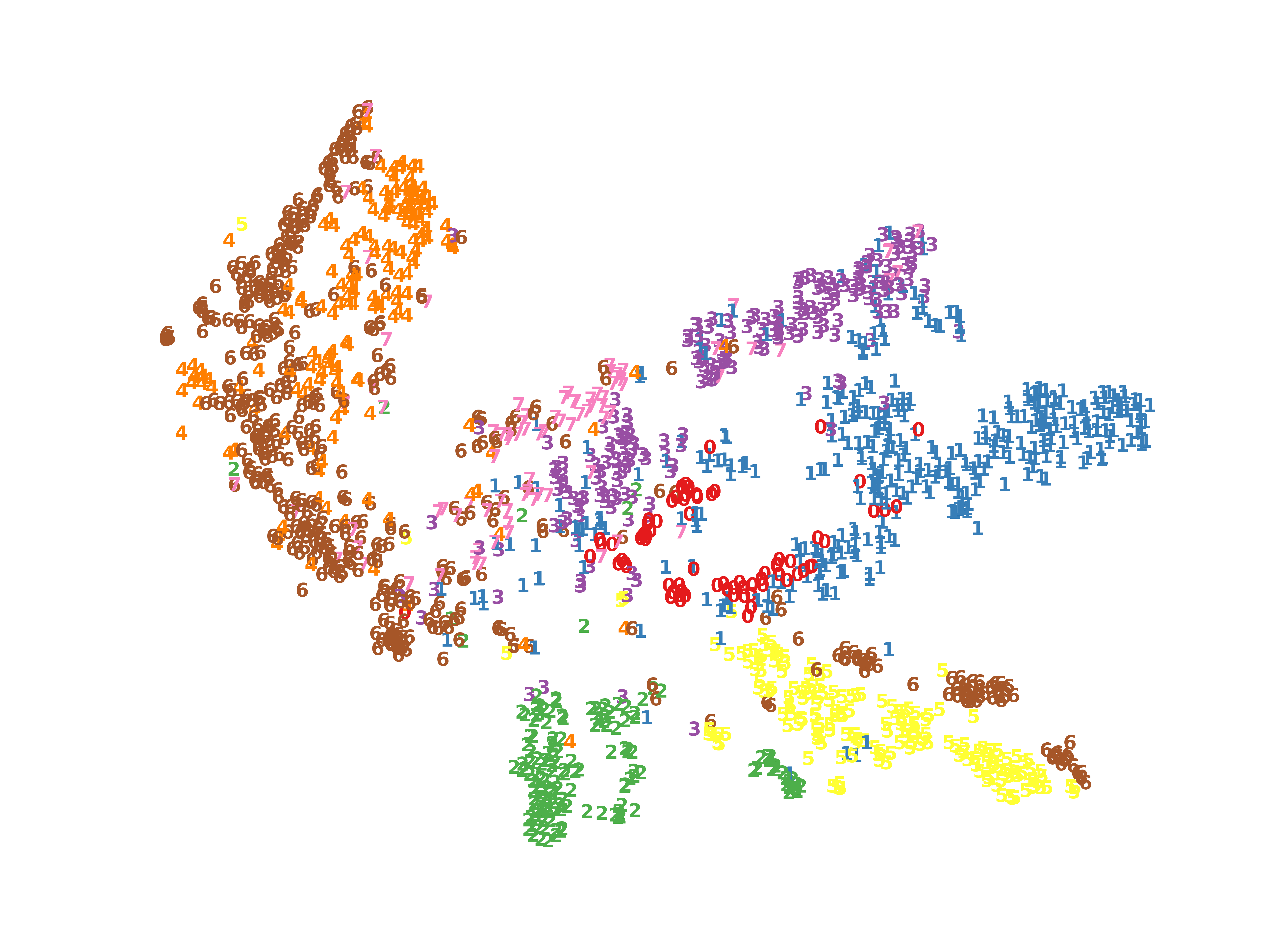}}
\vspace{3pt}
\centerline{(c) AutoSSL}
\end{minipage}
\begin{minipage}{0.139\linewidth}
\centerline{\includegraphics[width=\textwidth]{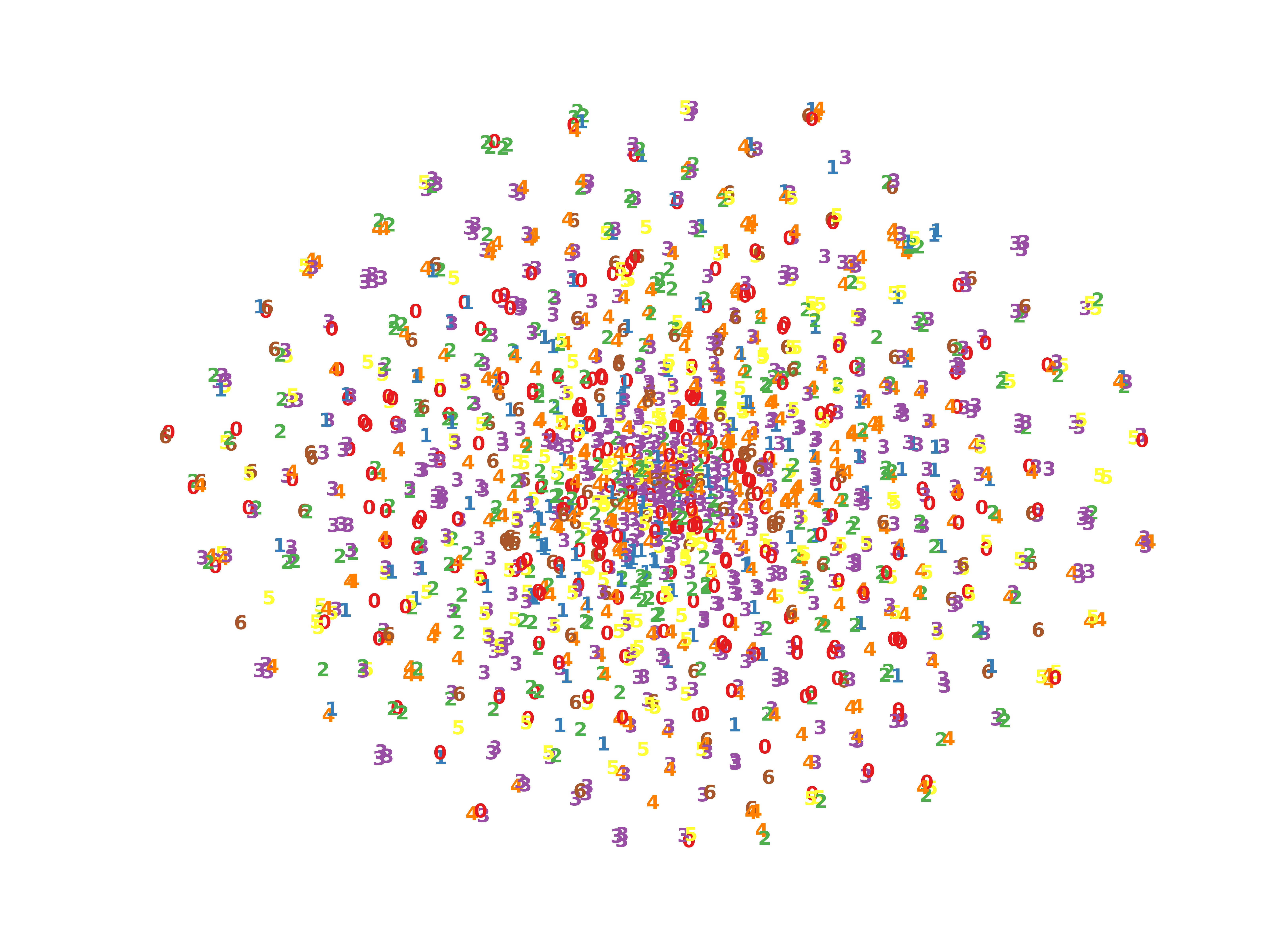}}
\vspace{3pt}
\centerline{\includegraphics[width=\textwidth]{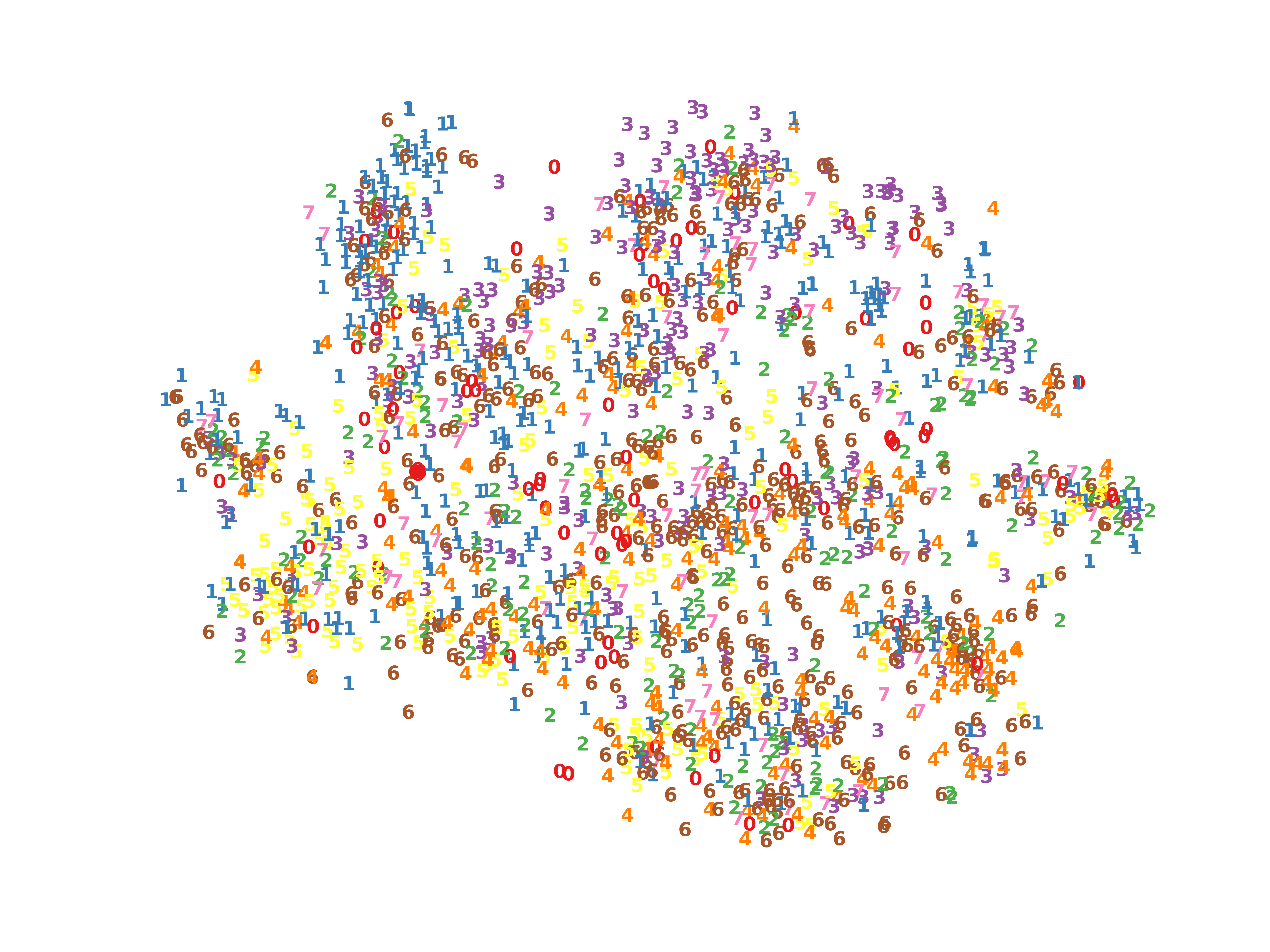}}
\vspace{3pt}
\centerline{(d) AFGRL}
\end{minipage}
\begin{minipage}{0.139\linewidth}
\centerline{\includegraphics[width=\textwidth]{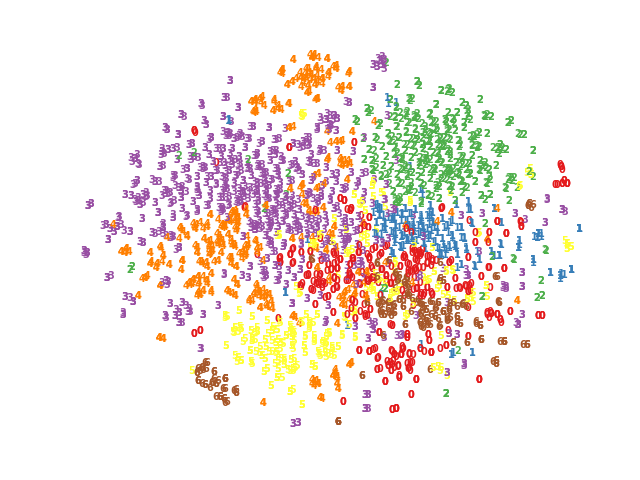}}
\vspace{3pt}
\centerline{\includegraphics[width=\textwidth]{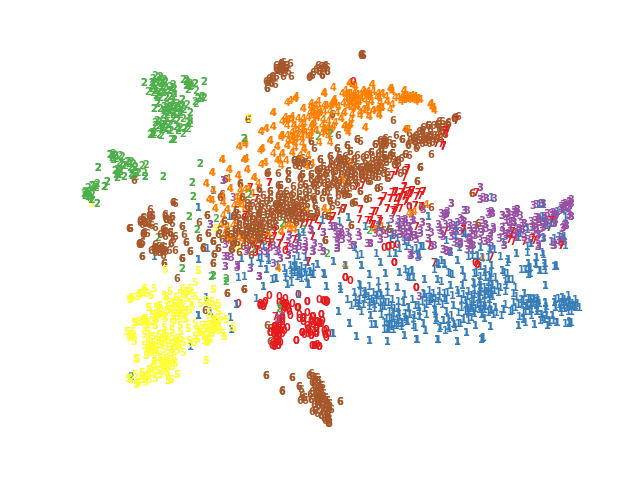}}
\vspace{3pt}
\centerline{(e) MVGRL}
\end{minipage}
\begin{minipage}{0.139\linewidth}
\centerline{\includegraphics[width=\textwidth]{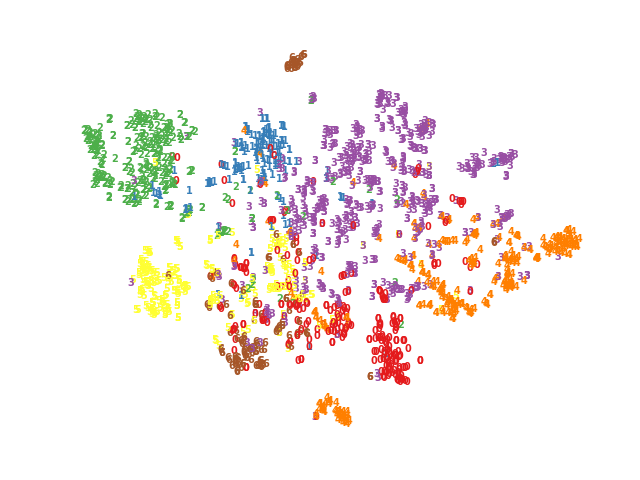}}
\vspace{3pt}
\centerline{\includegraphics[width=\textwidth]{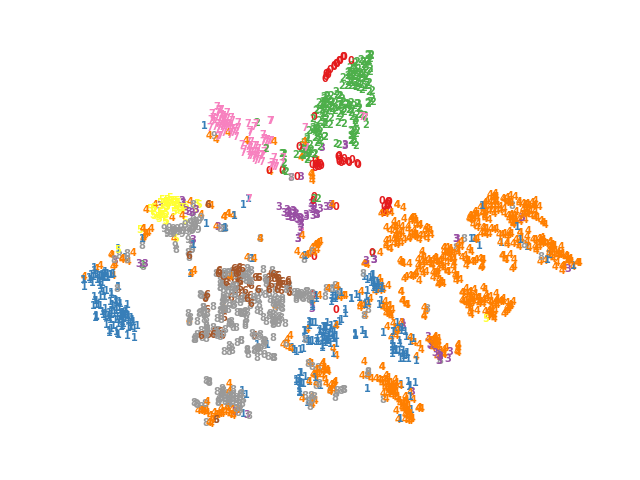}}
\vspace{3pt}
\centerline{(f) ProGCL}
\end{minipage}
\begin{minipage}{0.139\linewidth}
\centerline{\includegraphics[width=\textwidth]{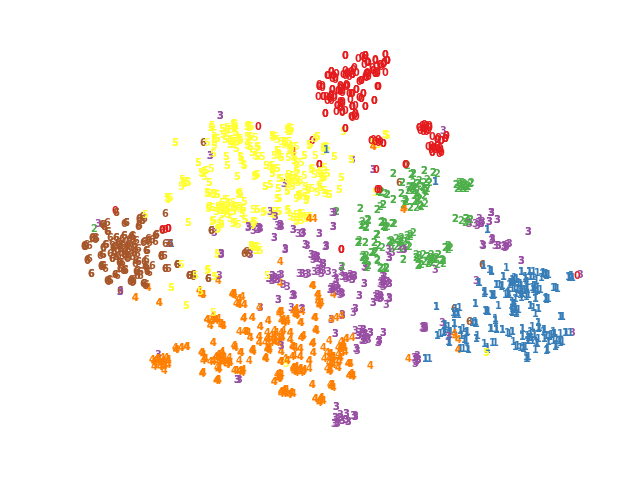}}
\vspace{3pt}
\centerline{\includegraphics[width=\textwidth]{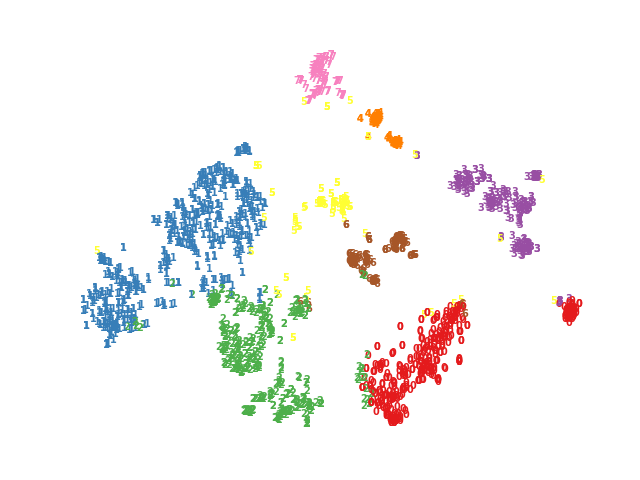}}
\vspace{3pt}
\centerline{(g) Ours}
\end{minipage}
\caption{2D $t$-SNE visualization of seven methods on two benchmark datasets. The first row and second row corresponds to CORA and AMAP dataset, respectively.}
\label{t_SNE}  
\end{figure*}

\subsection{Training \label{training}}
In this section, we introduce the training technique of the proposed RGC. It mainly contains two parts including encoding network training and quality network training. 

To train the encoder $\mathcal{F}$ in a self-supervised manner, we construct a contrastive pre-text task. Specifically, two-view node embeddings are generated and the network is forced to pull together the same nodes in different views while pushing away the different ones. The contrastive loss $\mathcal{L}_{con}$ is formulated as the infoNCE loss. In addition, a clustering guidance loss $\mathcal{L}_{clu}$ is adapted to align the clustering distribution with the sharpened ones. The detailed formulations of $\mathcal{L}_{con}$ and $\mathcal{L}_{clu}$ can be found in Appendix A.2. Thus, the encoder loss is formulated as $\mathcal{L}_{\mathcal{F}}=\mathcal{L}_{con}+\alpha \mathcal{L}_{clu}$, where $\alpha$ is the trade-off hyper-parameter.

For the $\mathcal{Q}$ network training, the experience replay training strategy is adopted. Concretely, during training, we will collect the experience quadruple tuples including states, actions, next states, and rewards into the experience buffer $\mathcal{B}$ as follow.
\begin{equation} 
\mathcal{B} = \{\textbf{S}_t,\hat{K}_t,\textbf{S}_{t+1},R_t|t\in[t_s,t_e)\},
\label{buffer}
\end{equation}
where $t_s$ and $t_e$ are the start epoch and the end epoch of the collection. Based on the collected experience buffer $\mathcal{B}$, the quality network $\mathcal{Q}$ will be trained by minimizing the following reinforcement learning loss:
\begin{equation} 
\begin{aligned}
\mathcal{L}_{\mathcal{Q}} = \frac{1}{t_e-t_s} \sum_{t=t_s}^{t_e-1} (R_t+\gamma \cdot \max \textbf{q}_{t+1} -\textbf{q}_t[\hat{K}_t])^2= \\ \frac{1}{t_e-t_s} \sum_{t=t_s}^{t_e-1} (R_t+\gamma \cdot \max \mathcal{Q}(\textbf{S}_{t+1}) - \mathcal{Q}(\textbf{S}_{t})[\hat{K}_t])^2,
\end{aligned}
\label{Q_loss}
\end{equation}
where $\gamma$ is a discount factor for the future step. In Eq. \eqref{Q_loss}, the term $R_t+\gamma \cdot max \mathcal{Q}(\textbf{S}_{t+1})$ is the quality estimation of action $\hat{K}_t$ at state $\textbf{S}_t$. Besides, the term $\mathcal{Q}(\textbf{S}_{t})[\hat{K}_t]$ is the learned quality of action $\hat{K}_t$ at state $\textbf{S}_t$. By minimizing $\mathcal{L}_{\mathcal{Q}}$, $\mathcal{Q}$ is guided to better learn the quality of actions and take intelligent action. 

In summary, the encoder network is trained by minimizing the encoder loss $\mathcal{L}_{\mathcal{F}}$ and the quality network $\mathcal{Q}$ is optimized by minimizing the reinforcement learning loss $\mathcal{L}_{\mathcal{Q}}$. By these settings, our proposed RGC is endowed with strong representation capability and cluster number recognition capability, thus achieving promising performance.

% \subsubsection{Experience Replay Buffer}

% \subsubsection{Loss Function \label{train_loss}}

\subsection{Complexity Analysis}
In this section, we analyze the time and space complexity of calculating the loss functions in our proposed RGC. Assume that the max cluster number, the experience buffer size, and the encoding time for one state is $N_K$, $N_{\mathcal{B}}$, and $\text{T}_\mathcal{F}$, respectively. Besides, one state memory cost, the sample number, and the latent feature dimensions denote $M_{\textbf{S}}$, $N$, and $d$, respectively. Thus, the time complexity of calculating $\mathcal{L}_{\mathcal{Q}}$, $\mathcal{L}_{con}$, $\mathcal{L}_{clu}$ is $\mathcal{O}(N_{\mathcal{B}}\text{T}_\mathcal{F}+N_{\mathcal{B}}N_K/2+N_{\mathcal{B}}/2)$, $\mathcal{O}(N^2d)$, and $\mathcal{O}(N\hat{K})$, respectively. In addition, the space complexity of $\mathcal{L}_{\mathcal{Q}}$, $\mathcal{L}_{con}$, $\mathcal{L}_{clu}$ is $\mathcal{O}(N_{\mathcal{B}}M_{\textbf{S}})$, $\mathcal{O}(N^2)$, and $\mathcal{O}(N\hat{K})$, respectively. The detailed calculation process can be found in Appendix A.4.

\section{Experiment}
To demonstrate the effectiveness of our proposed Reinforcement Graph Clustering (RGC), we conduct extensive experiments. In the following sections, we introduce the datasets and experimental setups first. And then, we conduct comparison experiments, ablation studies, and analysis experiments. 

\subsection{Dataset}
To comprehensively compare our proposed RGC with the state-of-the-art baselines, the experiments are conducted in five benchmark datasets, including CORA \cite{AGE}, CITESEER \cite{SDCN}, Amazon Photo (AMAP)\cite{DCRN}, Brazil Air-Traffic (BAT)\cite{R-GAE} and Europe Air-Traffic (EAT)\cite{R-GAE}. The detailed information of datasets are summarized in Table 1 of the Appendix.

% \ref{DATASET_INFO}

\subsection{Experimental Setup}
In our paper, all experiments are conducted on the desktop computer with the Intel Core i7-7820x CPU, one NVIDIA GeForce RTX 2080Ti GPU, 64GB RAM, and the PyTorch deep learning platform. For the compared state-of-the-art baselines, we adopt their source with original settings and reproduce the results. In our method, we train the encoder $\mathcal{F}$ for 400 epochs. Besides, the quality network $\mathcal{Q}$ is trained for 30 epochs whenever the experience buffer is full. The learning rate of $\mathcal{Q}$ is set to $1e^{-3}$ and the learning rate of $\mathcal{F}$ is selected from $\{1e^{-5}, 1e^{-4}, 1e^{-3}\}$. Besides, the experience buffer size is searched from $\{30,40,50\}$ and the initial greedy rate $\epsilon$ is selected from $\{0.3, 0.5, 0.7\}$. The initial greedy rate will increase linearly during training. The max cluster number $N_K$ and the trade-off parameter $\alpha$ is set to 10. Besides, the discount rate $\gamma$ is set to 0.1. For the clustering algorithm $\mathcal{P}$, we adopt $K$-Means and the clustering performance is evaluated by two widely-used metrics, i.e., NMI and ARI \cite{jiaqijin,caijinyu_1,caijinyu_2,caijinyu_3,huangsheng_2,huangsheng_1}.

% is set to 10 

\begin{table}[!t]
\caption{Clustering performance of the traditional parametric methods$^\circ$, traditional non-parametric methods$^\bullet$, and our proposed RGC$^\ddag$. All results are obtained by ten runs and reported with mean±std (\%). The bold and underlined result denotes the champion and runner-up.}
\scalebox{0.7}{
\begin{tabular}{cc|c|c|c|c|c}
\hline
\multicolumn{2}{c|}{\multirow{2}{*}{\textbf{Method}}}                  & \textbf{$K$-Means}$^\circ$  & \textbf{GMM}$^\circ$     & \textbf{DBSCAN}$^\bullet$     & \textbf{DPCA}$^\bullet$        & \textbf{RGC}$^\ddag$        \\ \cline{3-7} 
\multicolumn{2}{c|}{}                                                  & \cite{K-means}        & \cite{GMM}        & \cite{DBSCAN}           & \cite{DPCA}            & Ours           \\ \hline
\multicolumn{1}{c|}{\multirow{3}{*}{\textbf{CORA}}}     & \textbf{NMI} & 26.42+1.03       & 27.21+1.00       & {\underline{38.46+0.37}}    & 14.30+0.40           & \textbf{57.60+1.36} \\
\multicolumn{1}{c|}{}                                   & \textbf{ARI} & 12.97+3.32       & {\underline{15.03+1.02}} & 07.37+1.07          & 12.42+0.94           & \textbf{49.46+2.72} \\
\multicolumn{1}{c|}{}                                   & $K$   & -                & -                & 103.0+3.56          & {\underline{15.21+0.84}}     & \textbf{07.90+1.76} \\ \hline
\multicolumn{1}{c|}{\multirow{3}{*}{\textbf{CITESEER}}} & \textbf{NMI} & {\underline{34.70+0.66}} & 34.16+0.83       & 09.01+0.47          & 14.47+0.87           & \textbf{45.06+0.93} \\
\multicolumn{1}{c|}{}                                   & \textbf{ARI} & {\underline{30.72+1.35}} & 30.08+1.40       & 01.60+0.09          & 13.24+0.92           & \textbf{46.17+1.46} \\
\multicolumn{1}{c|}{}                                   & $K$   & -                & -                & {\underline{32.67+2.05}}    & 45.58+0.24           & \textbf{05.40+0.49} \\ \hline
\multicolumn{1}{c|}{\multirow{3}{*}{\textbf{AMAP}}}     & \textbf{NMI} & 18.98+1.83       & {\underline{19.74+1.79}} & 17.21+0.03          & \multirow{3}{*}{OOM} & \textbf{69.61+0.36} \\
\multicolumn{1}{c|}{}                                   & \textbf{ARI} & 07.83+0.51       & 07.96+0.68       & {\underline{08.95+0.01}}    &                      & \textbf{59.58+0.39} \\
\multicolumn{1}{c|}{}                                   & $K$   & -                & -                & {\underline{7.33+0.47}}     &                      & \textbf{09.40+0.66} \\ \hline
\multicolumn{1}{c|}{\multirow{3}{*}{\textbf{BAT}}}      & \textbf{NMI} & 30.50+1.66       & 30.50+0.00       & {\underline{43.53+0.00}}    & 20.92+0.04           & \textbf{51.58+0.83} \\ 
\multicolumn{1}{c|}{}                                   & \textbf{ARI} & 12.71+0.74       & 12.71+0.00       & {\underline{36.12+0.02}}    & 22.82+0.01           & \textbf{47.16+1.35} \\ 
\multicolumn{1}{c|}{}                                   & $K$   & -                & -                & {\underline{03.33+0.47}}    & 16.02+0.02           & \textbf{03.80+0.60} \\ \hline

% \multicolumn{1}{c|}{\multirow{3}{*}{\textbf{EAT}}}      & \textbf{NMI} & 30.50+1.66       & 30.50+0.00       & {\underline{43.53+0.00}}    & 20.92+0.04           & \textbf{51.58+0.83} \\ 
% \multicolumn{1}{c|}{}                                   & \textbf{ARI} & 12.71+0.74       & 12.71+0.00       & {\underline{36.12+0.02}}    & 22.82+0.01           & \textbf{47.16+1.35} \\ 
% \multicolumn{1}{c|}{}                                   & $K$   & -                & -                & {\underline{03.33+0.47}}    & 16.02+0.02           & \textbf{03.80+0.60} \\ \hline

\multicolumn{1}{c|}{\multirow{3}{*}{\textbf{EAT}}}      & \textbf{NMI} & {\underline{13.52+1.19}} & 13.52+0.00       & 09.23+0.40          & 06.14+0.57           & \textbf{37.77+0.13} \\
\multicolumn{1}{c|}{}                                   & \textbf{ARI} & 02.79+1.54       & 02.79+0.00       & {\underline{12.39+0.35}}    & 07.58+0.26           & \textbf{30.16+0.15} \\ 
\multicolumn{1}{c|}{}                                   & $K$   & -                & -                & \textbf{04.50+0.24} & 08.50+0.54           & {\underline{03.30+0.30}}    \\ \hline
\end{tabular}
}
\label{compare_traditional}
\end{table}

\subsection{Clustering Performance Comparison}
In this section, we conduct extensive performance comparison experiments to demonstrate the effectiveness of our proposed method.

As shown in Table \ref{compare}, we first compare our RGC with twenty deep parametric state-of-the-art baselines. Concretely, DEC \cite{DEC}, DCN \cite{AE_K_MEANS}, IDEC \cite{IDEC}, and AdaGAE \cite{AdaGAE} are four representative deep clustering methods. Besides, MGAE \cite{MGAE}, DAEGC \cite{DAEGC}, ARGA \cite{ARGA_conf}, SDCN \cite{SDCN}, DFCN \cite{DFCN} are five generative deep graph clustering methods, which reconstruct the attribute or/and topological information. Moreover, AGE \cite{AGE}, MVGRL \cite{MVGRL}, GDCL \cite{GDCL}, GCA \cite{GCA}, MCGC \cite{MCGC}, AutoSSL \cite{AutoSSL}, AGC-DRR \cite{AGC-DRR}, DCRN \cite{DCRN}, AFGRL \cite{AFGRL}, ProGCL \cite{ProGCL}, and SUBLIME \cite{SUBLIME} are eleven contrastive deep graph clustering methods. They improve the discriminative capability of samples by pulling together the positive samples while pushing away the negative ones. However, previous methods are parametric, and their promising performance heavily relies on the predefined cluster number $K$, which is not always available in the real scenario. To solve this problem, we propose RGC by determining the cluster number automatically with reinforcement learning. From the results in Table \ref{compare}, we have three observations. 1) The deep clustering methods can not achieve promising performance since they merely learn the attribute information while neglecting the graph topological information. 2) Our proposed RGC achieves better performance than the generative methods. The reason is that we adopt the contrastive mechanism, thus improving the discriminative capability of samples. 3) RGC is comparable with the contrastive methods. Here, actually, the comparison is ``not fair'' since we provide the correct cluster number for the contrastive methods and not for our method RGC. Even under unfair conditions, our proposed RGC is still comparable with state-of-the-art deep parametric methods. The main reason is that the proposed cluster number learning module automatically estimates the cluster number, and the encoder learns the discriminative node embeddings, thus achieving promising performance.

In addition, we also compare our proposed RGC with traditional parametric methods and traditional non-parametric methods. To be specific, two representative traditional parametric methods, $K$-Means \cite{K-means} and GMM \cite{GMM}, adopt the idea of exception maximum to perform clustering. But they still rely on the predefined cluster number. Differently, two traditional non-parameter algorithms DBSCAN \cite{DBSCAN} and DPCA \cite{DPCA}, are free from the predefined cluster number. As shown in Table \ref{compare_traditional}, we have two conclusions as follows. 1) The traditional parametric methods can not achieve promising performance since the representation capability is weak compared with our method. 2) Although the traditional non-parametric methods, including DBSCAN \cite{DBSCAN} and DPCA \cite{DPCA}, are free from cluster numbers, they are not comparable with ours. The reasons are as follows. Firstly, their representation capability is weaker than the deep methods. Secondly, they are merely designed for the non-graph data and neglect the structural information in graphs. 

% 2) Besides, the traditional parametric methods like $K$-Means \cite{K-means} and GMM \cite{GMM} still rely on the pre-defined cluster number $K$. 

In summary, our proposed RGC can outperform the traditional parametric/non-parametric methods and achieve comparable performance compared with state-of-the-art deep parametric methods. In the next section, we conduct time cost comparison experiments of our proposed cluster number learning method with the existing cluster number estimation methods.

% including DBSCAN and . 

\begin{figure}[h]
\centering
\includegraphics[scale=0.50]{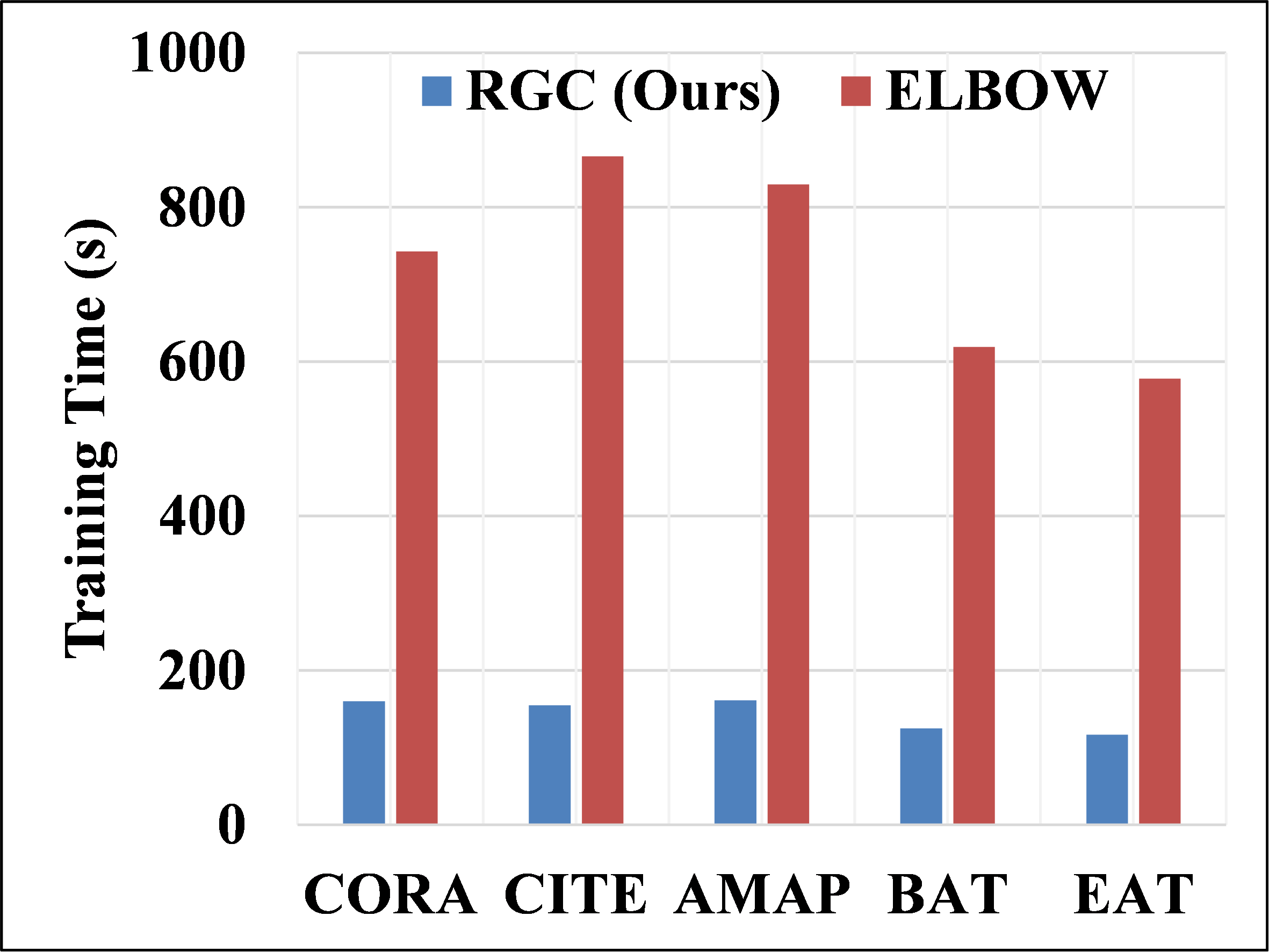}
\caption{Training time comparison of our proposed RGC and the $K$ estimation method ELBOW applied to deep graph clustering.}
\label{time_cost}  
\end{figure}

\subsection{Time Cost Comparison of $K$ Estimation\label{time_efficiency}} 
The existing deep graph clustering methods are parametric, i.e., relying on the predefined cluster number. In the field of traditional clustering, some cluster number estimation methods like ELBOW \cite{ELBOW} can help to determine the cluster numbers. However, it brings large computational costs to the deep graph clustering methods since the deep neural networks need to be trained repeatedly. To verify this, we conduct experiments by comparing the training time of our RGC with that of the $K$ estimation method. Concretely, for the $K$ estimation method, we remove the cluster number learning module in RGC and attempt to train the networks with cluster numbers from 2 to $N_K$. In Figure \ref{time_cost}, the training time of RGC and $K$ estimation method are visualized. From these results, we conclude that our method is more time efficient than the cluster number estimation method. The main reason is that the $K$ estimation method requires training the networks repeatedly with different cluster numbers for calculating the unsupervised criterion, such as WSS (within-cluster sum of squares). Only based on the unsupervised criterion calculated by different trained models the $K$ estimation methods like ELBOW can help to determine the cluster number. Different from them, our cluster number learning module will recognize the cluster number by reinforcement learning. Concretely, the experiences are collected in the process of representation learning and clustering to train the quality network. After optimization, the quality network can automatically determine the cluster number. In these settings, the deep neural networks will be trained only once in the unified framework of unsupervised representation learning and cluster number determination, thus saving training time. 

%  and our RGC can achieve promising clustering performance
% Besides, the cluster number determination and unsupervised representation learning are unified to the uniform framework by the reinforcement learning. 

% keep the same experimental settings and 

% 1) they will fail. 2) 

% Besides, we compare the time costs of our methods and the ELBOW methods on .

\begin{figure}[h]
\centering
\small
\begin{minipage}{0.48\linewidth}
\centerline{\includegraphics[width=0.9\textwidth]{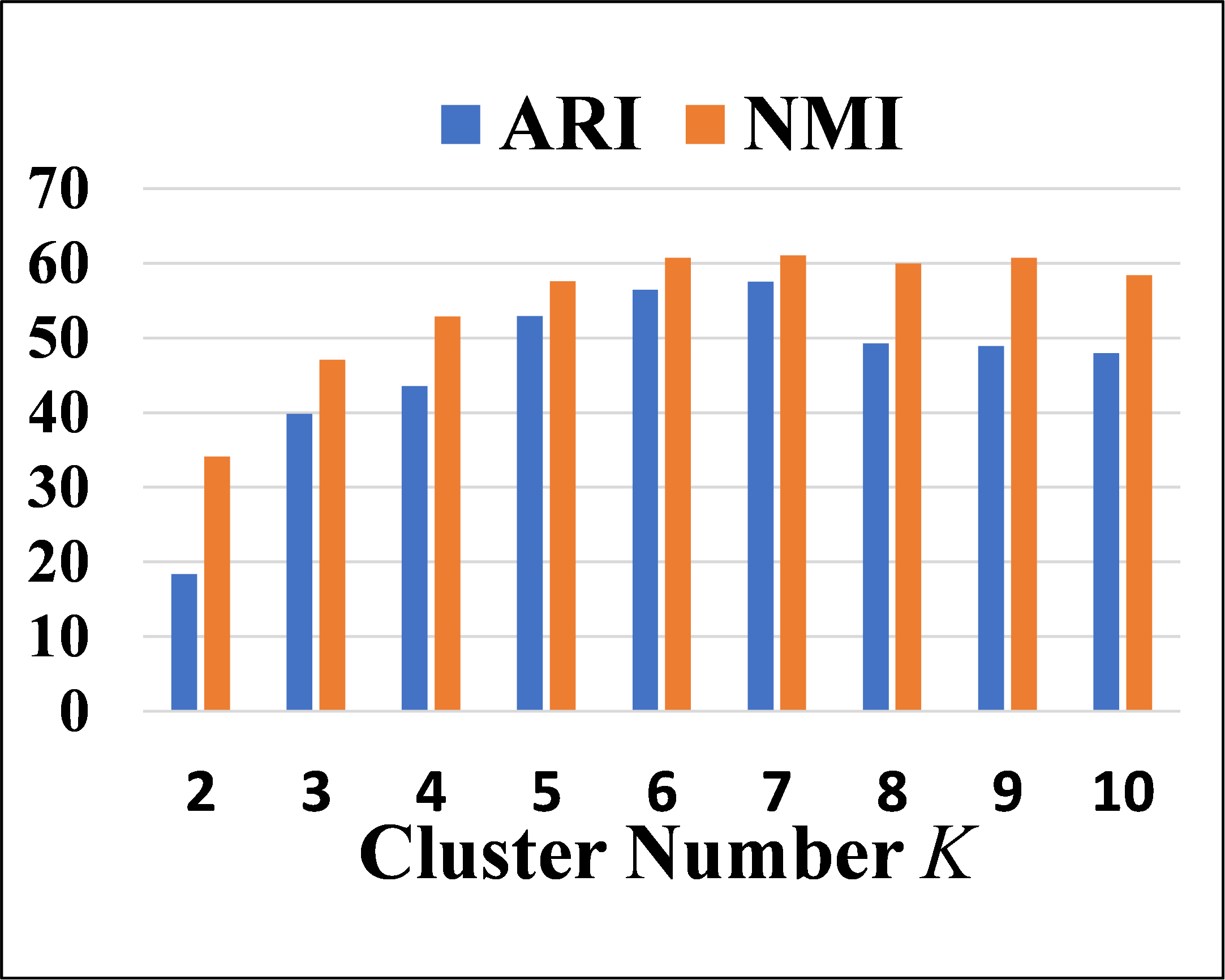}}
\centerline{(a) CORA}
\centerline{\includegraphics[width=0.9\textwidth]{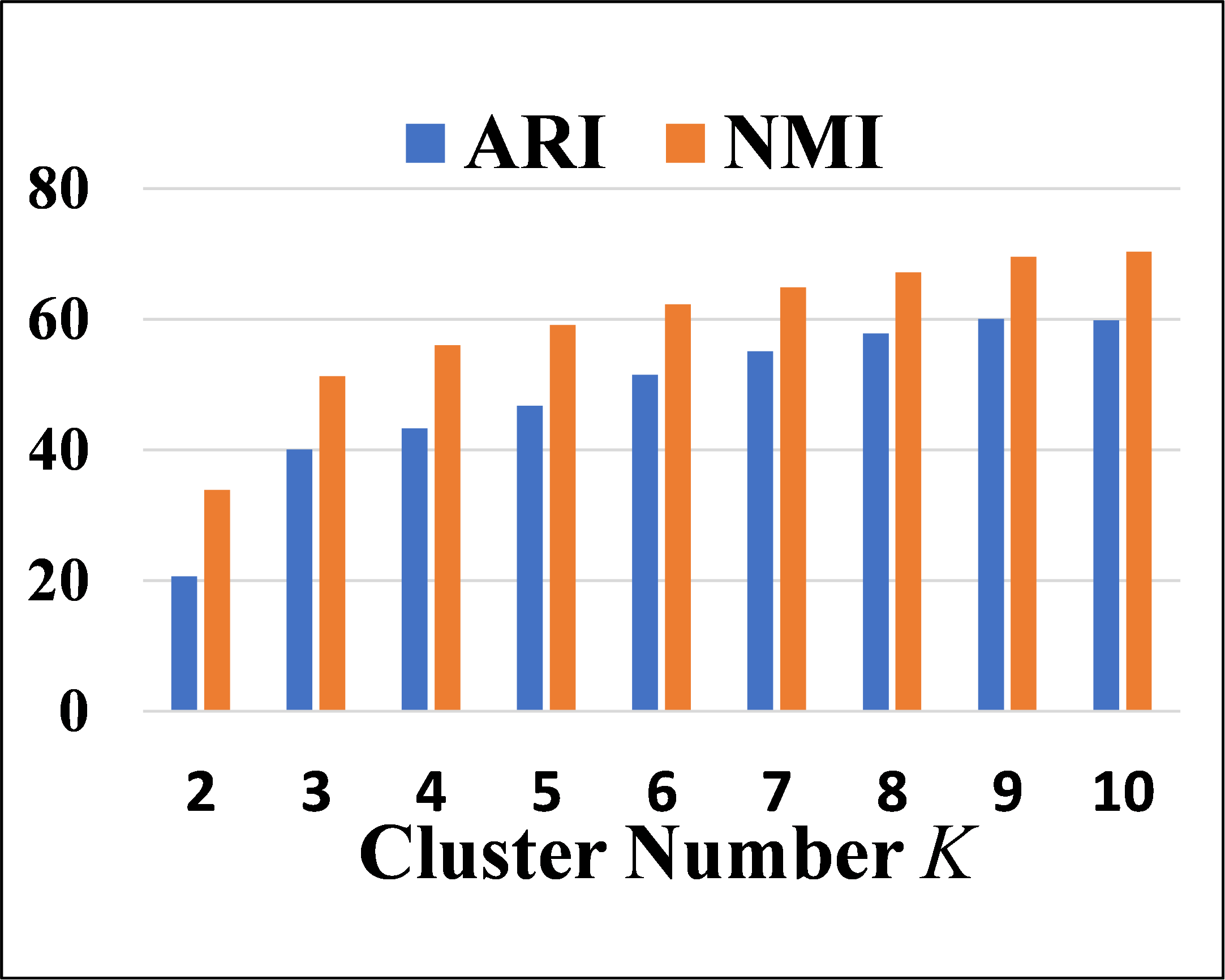}}
\centerline{(c) AMAP}
\end{minipage}
\begin{minipage}{0.48\linewidth}
\centerline{\includegraphics[width=0.9\textwidth]{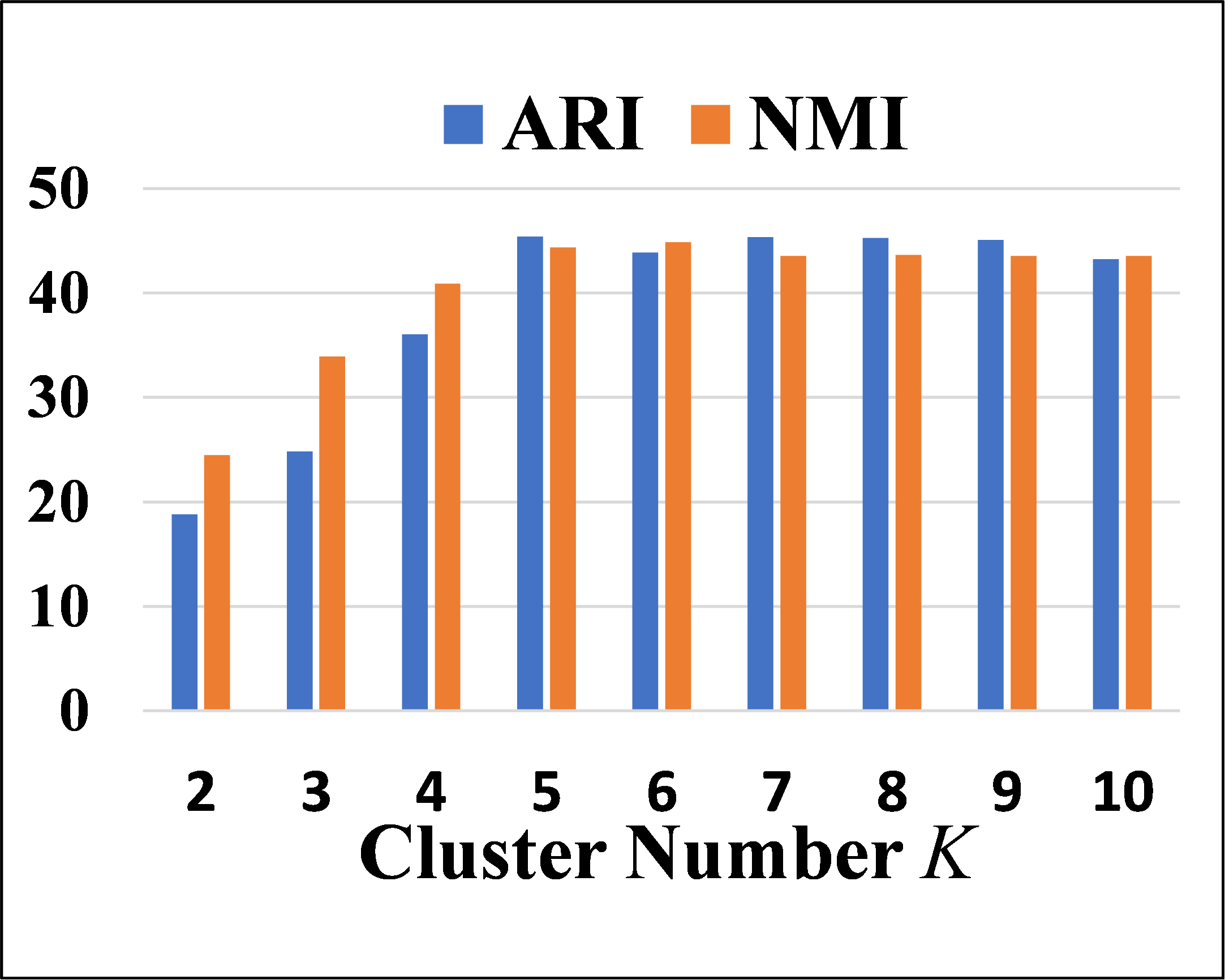}}
\centerline{(b) CITESEER}
\centerline{\includegraphics[width=0.9\textwidth]{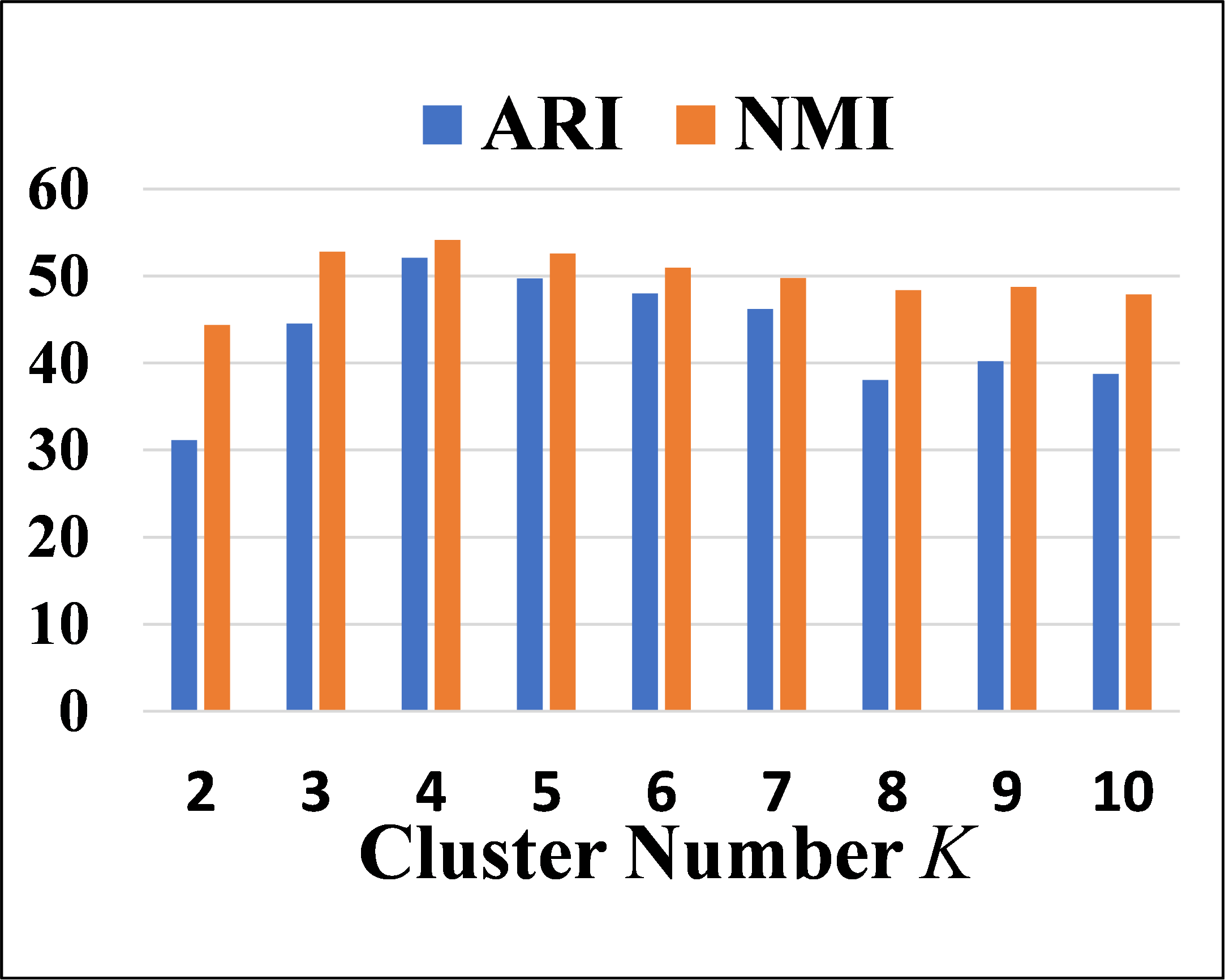}}
\centerline{(d) BAT}
\end{minipage}
% \begin{minipage}{0.48\linewidth}
% \centerline{\includegraphics[width=0.9\textwidth]{K_eat.png}}
% \centerline{(E) EAT}
% \end{minipage}
\caption{Clustering performance of the baseline trained with different cluster numbers $K$ on four datasets.}
\label{K_analysis}
\end{figure}

\subsection{Effectiveness of Learning Cluster Number}
To demonstrate the effectiveness of our proposed cluster number learning module, extensive experiments are conducted. 

At first, we demonstrate that the incorrect cluster number will lead to an unpromising performance in deep graph clustering. Concretely, we adopt our RGC without the cluster number learning module as the baseline. Then, the baseline is trained with different cluster numbers from 2 to $N_K$. In Figure \ref{K_analysis}, from these results, we have three conclusions as follows. 1) The wrong cluster number will lead to an unpromising performance in the parametric deep graph clustering methods. Take the results on the CORA dataset, for example, when $K$ is wrongly assigned to 2, the NMI will rapidly decrease to 34.10\%. But, when cluster number $K$ is 7, the clustering performance NMI can reach 61.05\%. 2) The clustering performance is sensitive to different cluster numbers. Take the results on AMAP as an example, the best NMI is 70.40\% while the worst NMI is 33.87\%. Besides, the standard deviation of NMI with different cluster numbers is 10.81\%. 3) The performance will reach the peak around the correct cluster number. But it might not reach the peak exactly at the correct cluster number. For example, on the CORA dataset, the performance reaches a peak at the correct cluster number 7. But on AMAP datasets, NMI reaches a peak at 10, which is around the ``correct'' cluster number 8. We hope this interesting phenomenon motivates the researchers to discover new classes or design algorithms to find the truly correct cluster number.

\begin{figure}[h]
\centering
\small
\begin{minipage}{0.48\linewidth}
\centerline{\includegraphics[width=0.9\textwidth]{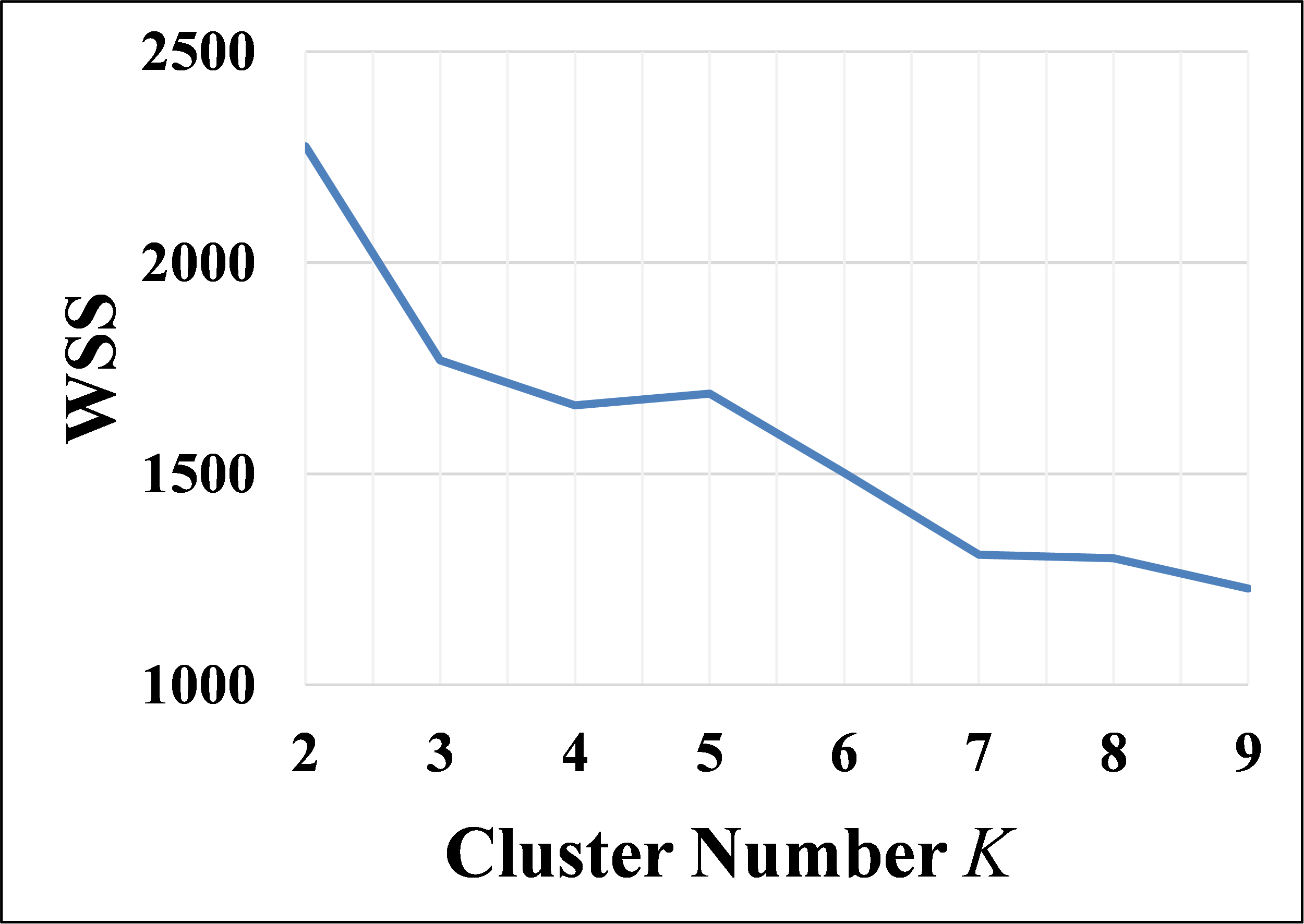}}
\centerline{(a) CORA}
\end{minipage}
\begin{minipage}{0.48\linewidth}
\centerline{\includegraphics[width=0.9\textwidth]{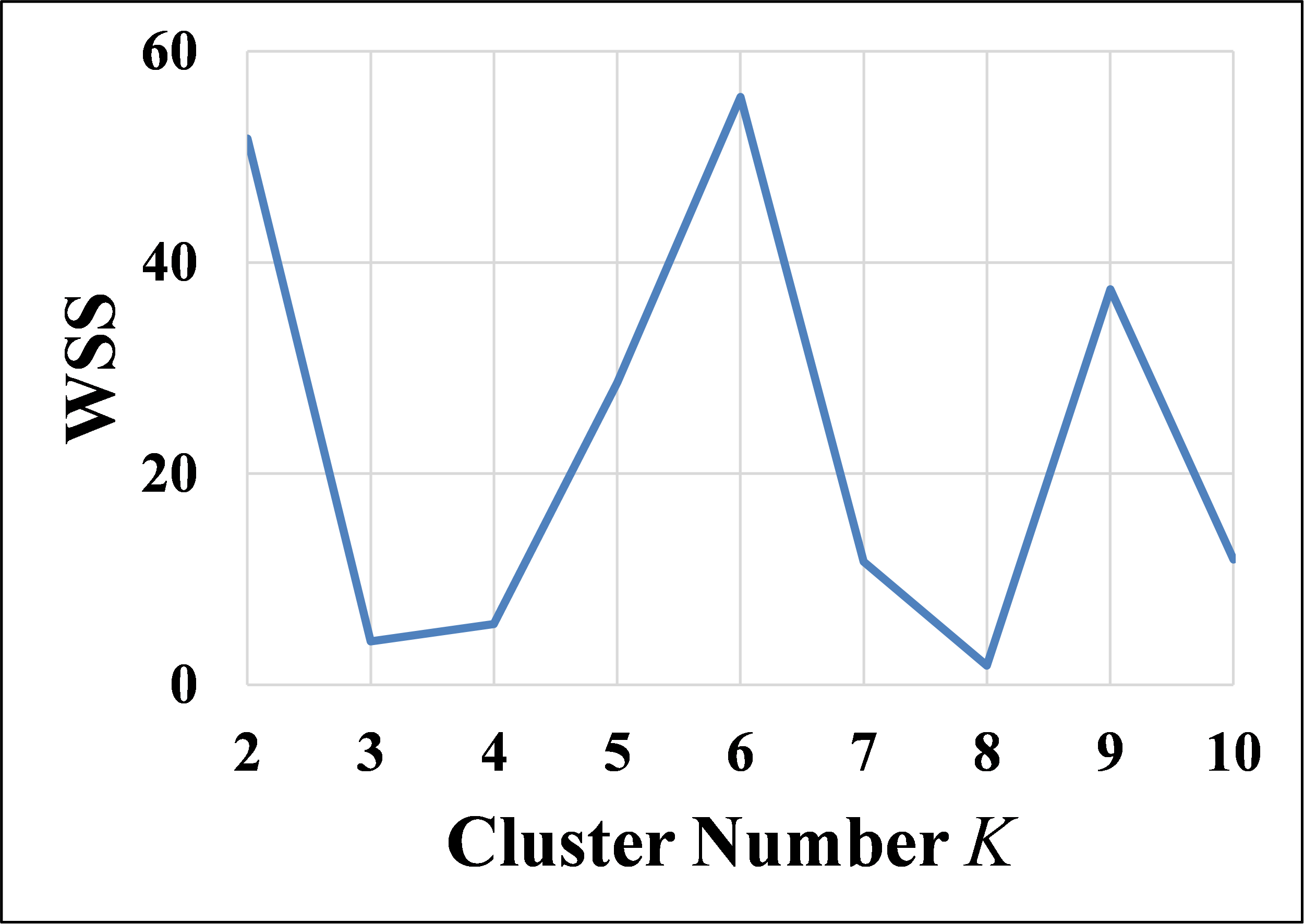}}
\centerline{(b) AMAP}
\end{minipage}
\caption{Determine cluster number with ELBOW \cite{ELBOW} cluster number estimation method.}
\label{ELBOW}
\end{figure}

Then, we adopt the cluster number estimation method ELBOW \cite{ELBOW} to determine the cluster number on CORA and AMAP datasets. Concretely, the baseline is RGC without the cluster number learning module. Besides, the baseline is trained with the cluster number from 2 to $N_K$, and the WSS (within-cluster sum of squares) is calculated to determine the cluster number. From the visualization results in Figure \ref{ELBOW}, we have three conclusions as follows. 1) The cluster determination based on the unsupervised criterion WSS needs professional experience. For example, on the CORA dataset, we observe that WSS first decrease dramatically when the cluster number equals three and then reaches a plateau. Thus, the cluster number might be wrongly assigned to 3 while the correct cluster number is 7. 2) Besides, the ELBOW method might fail when the trend of WSS is irregular. For example, on the AMAP dataset, the trend of WSS is irregular and we can not find any plateau, leading to the failure of ELBOW.

Different from the cluster number estimation method, our proposed cluster number learning module can automatically determine the cluster number by the reinforcement learning mechanism. In Table \ref{compare_traditional}, we observe that our RGC can recognize the cluster number precisely. For example, on the BAT dataset, the average of our learned cluster number is 3.80 under ten runs, and the ground truth is 4.

% Differerntly xx

% but not at the exactly correct cluster number.

% Besides, the trend of performance with different cluster numbers is irregular. For example, as $K$ increases, the performance also increases on AMAP dataset. However, on EAT

% remove the cluster number learning module in RGC and attempt to train the network with cluster numbers from 2 to $N_K$. As shown in Figure xxx, we have three observations as follows. 1)2)3) 

% we design and conduct

% \begin{figure}[h]
% \centering
% \small
% \begin{minipage}{0.48\linewidth}
% \centerline{\includegraphics[width=0.9\textwidth]{epsilon_cora.png}}
% \centerline{(a) CORA}
% \centerline{\includegraphics[width=0.9\textwidth]{epsilon_amap.png}}
% \centerline{(c) AMAP}
% \end{minipage}
% \begin{minipage}{0.48\linewidth}
% \centerline{\includegraphics[width=0.9\textwidth]{epsilon_cite.png}}
% \centerline{(b) CITESEER}
% \centerline{\includegraphics[width=0.9\textwidth]{epsilon_bat.png}}
% \centerline{(d) BAT}
% \end{minipage}
% % \begin{minipage}{0.48\linewidth}
% % \centerline{\includegraphics[width=0.9\textwidth]{epsilon_eat.png}}
% % \centerline{(e) EAT}
% % \end{minipage}
% \caption{Sensitivity analysis of the initial greedy rate hyper-parameter $\epsilon$ on four datasets.}
% \label{epsilon}
% \end{figure}

\subsection{Analysis}
In this section, we conduct analysis experiments for our proposed RGC. To be specific, we analyze the hyper-parameter, $t$-SNE visualization, and loss convergence in the following sub-sections.

\subsubsection{Hyper-parameter Analyses}
Due to the page limitation, we conduct hyper-parameters analyses in Appendix. 
% We analyze the sensitivity of hyper-parameters, including $\epsilon$ and $\alpha$ in our proposed RGC. Specifically, $\epsilon$ is the initial greedy rate and $\alpha$ denotes the trade-off parameter in $\mathcal{L}_{\mathcal{F}}$. As shown in Figure \ref{epsilon}, we conduct the sensitivity analysis experiments of hyper-parameter $\epsilon$ and have two conclusions as follows. 1) Our proposed RGC is not sensitive to the initial greedy rate $\epsilon$. It indicates that our method is not sensitive to initialization and will be optimized to achieve good clustering performance. 2) RGC can achieve promising performance when $\epsilon \in [0.3, 0.7]$. Thus, in our method, we search $\epsilon$ in $\{0.3,0.5,0.7\}$. Besides, as shown in Figure \ref{alpha}, we analyze the trade-off parameter $\alpha$ on four datasets and have two findings as follows. 1) Our proposed method is not sensitive to the trade-off parameter $\alpha$ when $\alpha \in [0.1, 10]$. 2) RGC will achieve promising performance on CITESEER, AMAP, and BAT datasets when $\alpha=10$. Thus, in our proposed method, $\alpha$ is set to 10. Additional Experimental results can be found in Figure \ref{eat_hyper} of the Appendix.

\subsubsection{$t$-SNE Visualization Analysis}
% Due to the page limitation, visualization 
We visualize the samples in the latent space by $t$-SNE algorithm \cite{T_SNE}. As shown in Figure \ref{t_SNE}, the experiments of seven compared methods are conducted on CORA and AMAP datasets. From the visualization results, our proposed RGC can better learn the clustering structure compared with other methods.

\begin{figure}[h]
\centering
\small
\begin{minipage}{0.48\linewidth}
\centerline{\includegraphics[width=0.9\textwidth]{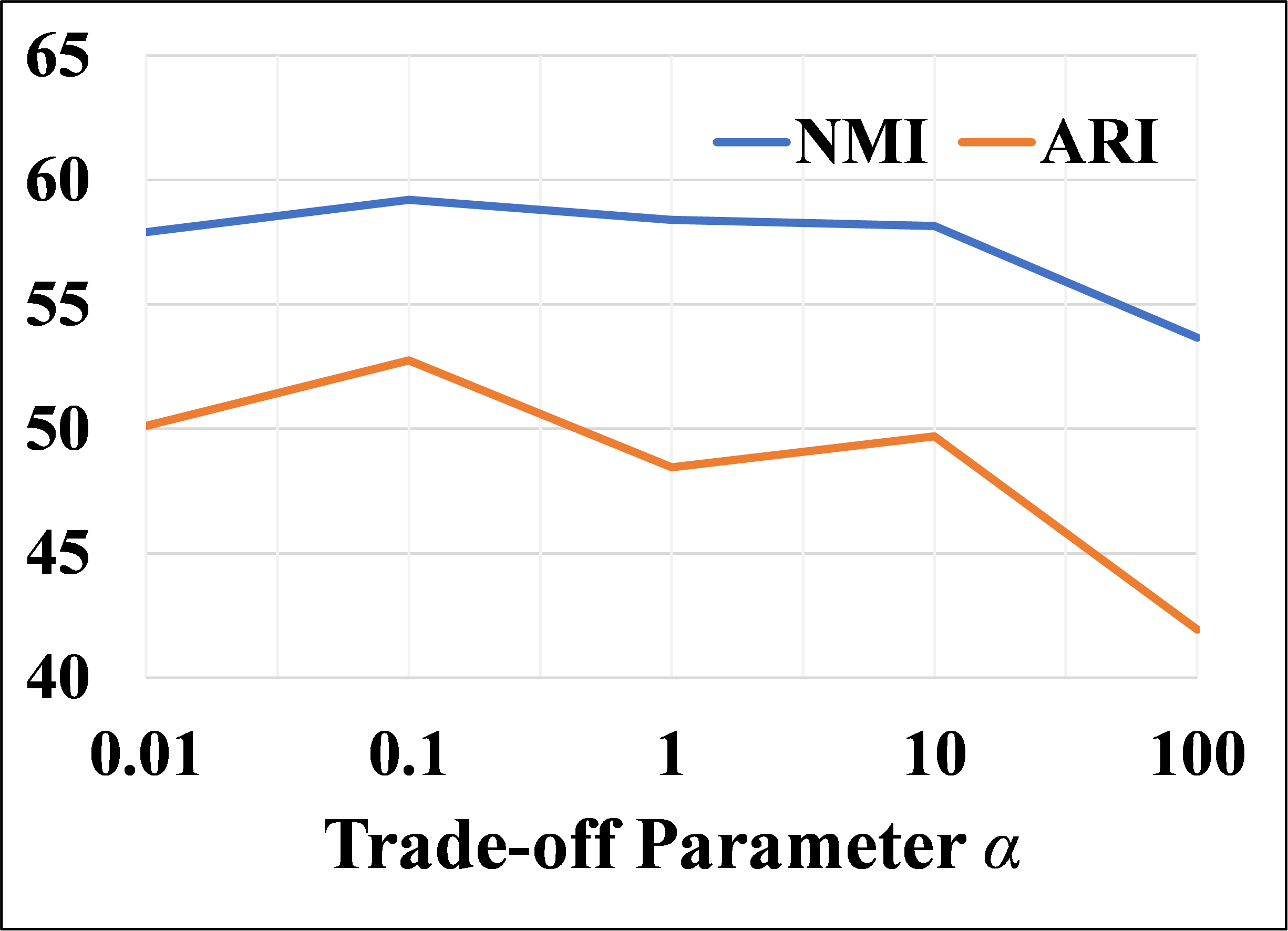}}
\centerline{(a) CORA}
\centerline{\includegraphics[width=0.9\textwidth]{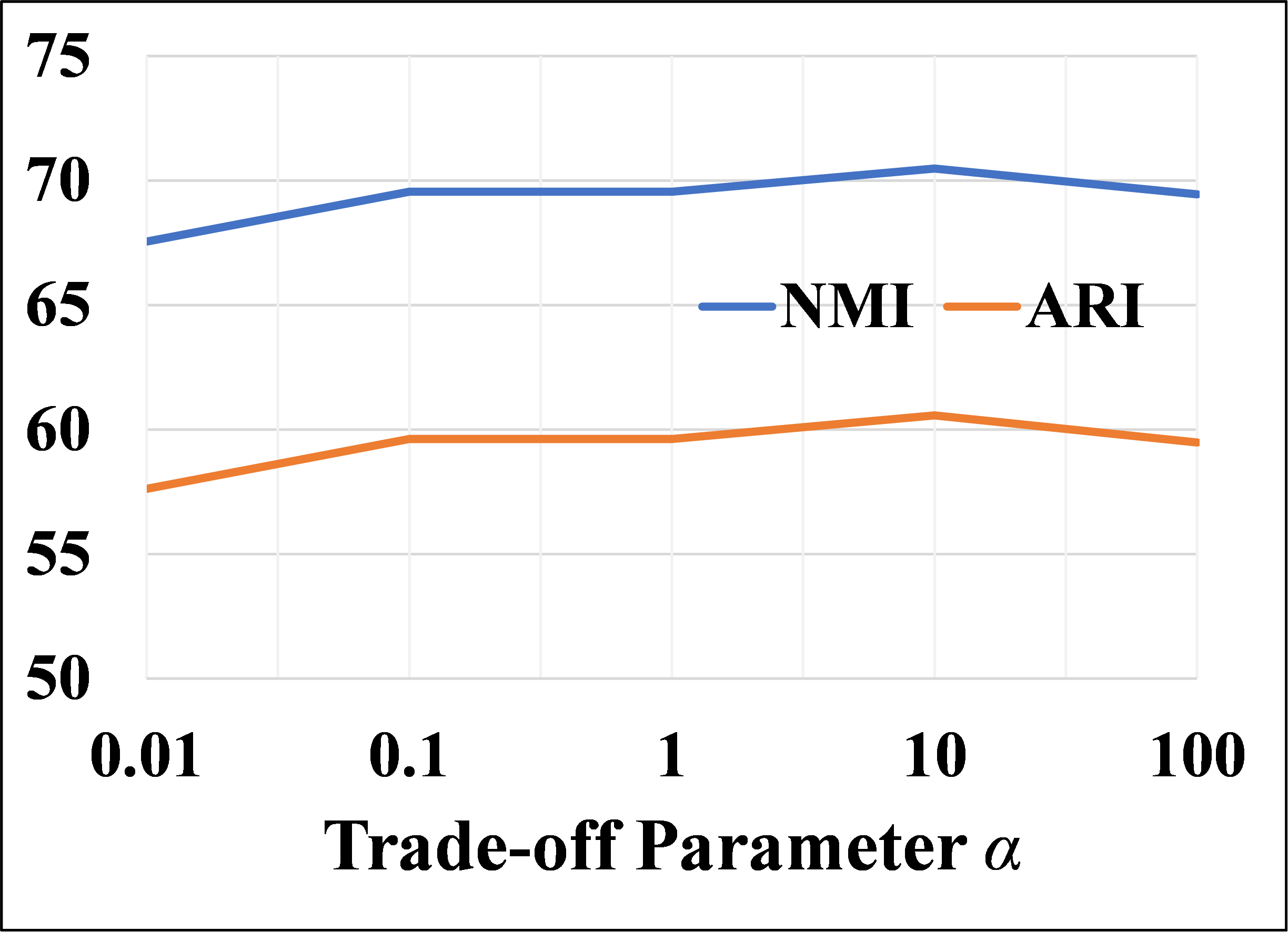}}
\centerline{(c) AMAP}
\end{minipage}
\begin{minipage}{0.48\linewidth}
\centerline{\includegraphics[width=0.9\textwidth]{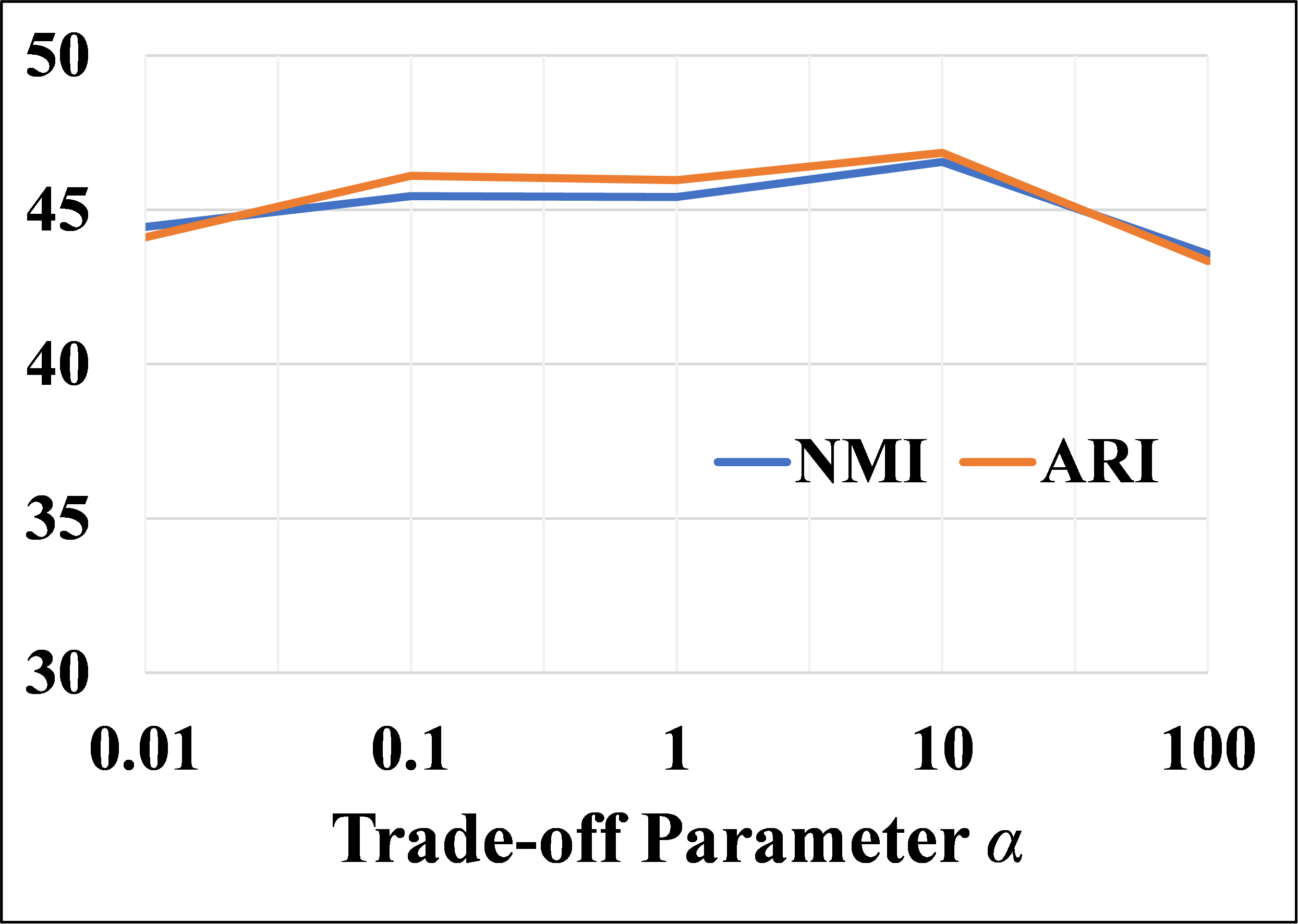}}
\centerline{(b) CITESEER}
\centerline{\includegraphics[width=0.9\textwidth]{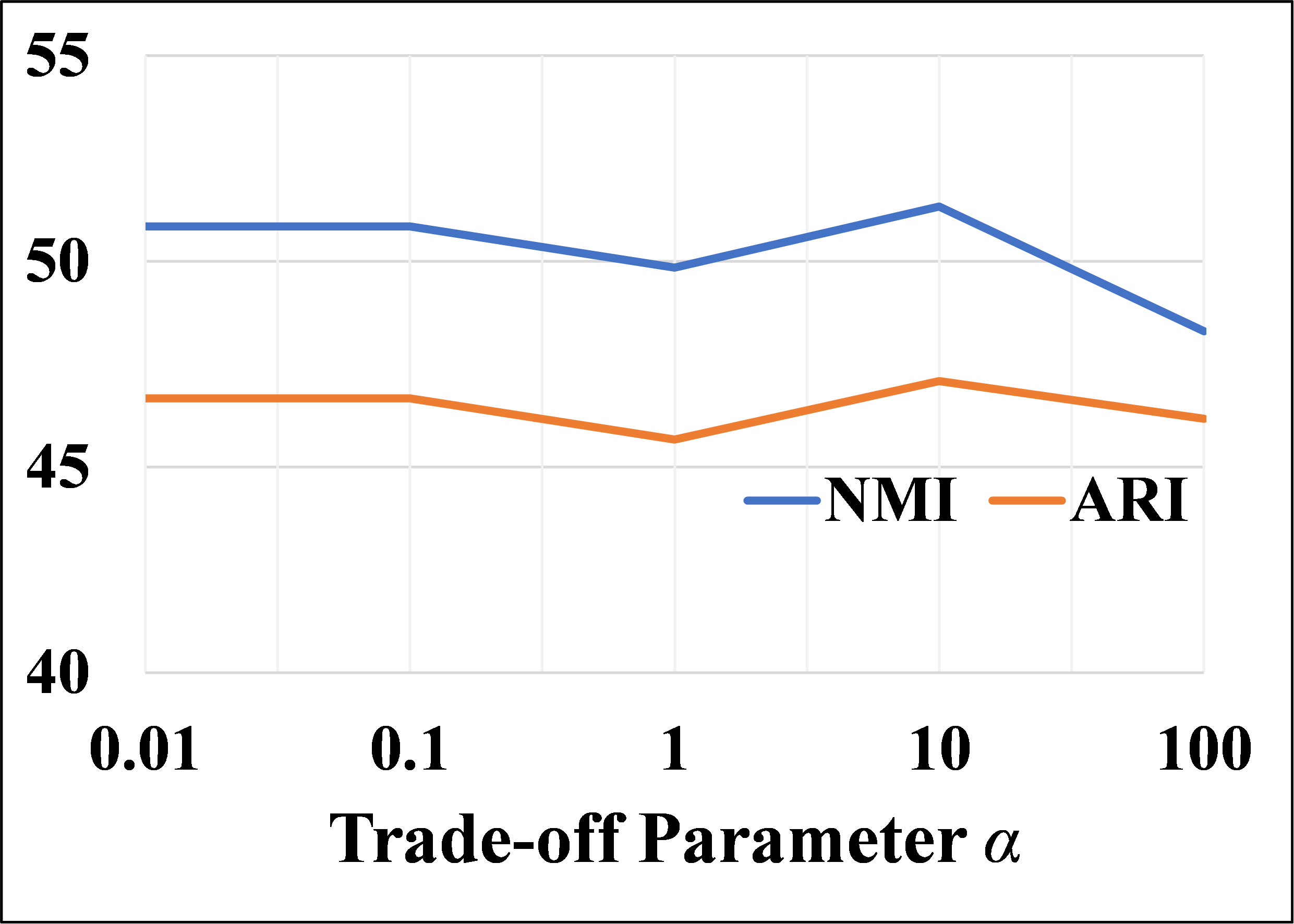}}
\centerline{(d) BAT}
\end{minipage}
% \begin{minipage}{0.48\linewidth}
% \centerline{\includegraphics[width=0.9\textwidth]{alpha_eat.png}}
% \centerline{(e) EAT}
% \end{minipage}
\caption{Hyper-parameter analysis of the trade-off parameter $\alpha$ on four datasets.}
\label{alpha}
\end{figure}

\subsubsection{Loss Convergence Analyses}
Due to the limitation of the main text, the loss convergence analyses are demonstrated in Appendix. 
% The loss functions of our proposed RGC include two parts, i.e., the encoder loss $\mathcal{L}_{\mathcal{F}}$ for training encoder $\mathcal{F}$ and the reinforcement loss $\mathcal{L}_{\mathcal{Q}}$ for training the quality network $\mathcal{Q}$. In this section, we plot the curve of loss functions on two datasets, CORA and BAT, as shown in Figure \ref{loss_function}. From these visualization results, the conclusion is that both $\mathcal{L}_{\mathcal{F}}$ and $\mathcal{L}_{\mathcal{Q}}$ can decrease during training and eventually converge.

\begin{figure}[h]
\centering
\small
\begin{minipage}{0.49\linewidth}
\centerline{\includegraphics[width=0.9\textwidth]{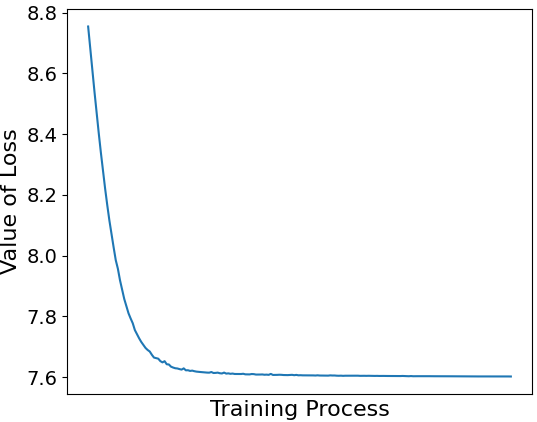}}
\centerline{(a) Encoder loss $\mathcal{L}_{\mathcal{F}}$ on CORA}
\vspace{3pt}
% \centerline{\includegraphics[width=0.9\textwidth]{CORA_Q_loss.png}}
% \centerline{(c) RL loss $\mathcal{L}_{\mathcal{Q}}$ on CORA}
\end{minipage}
\begin{minipage}{0.49\linewidth}
\centerline{\includegraphics[width=0.9\textwidth]{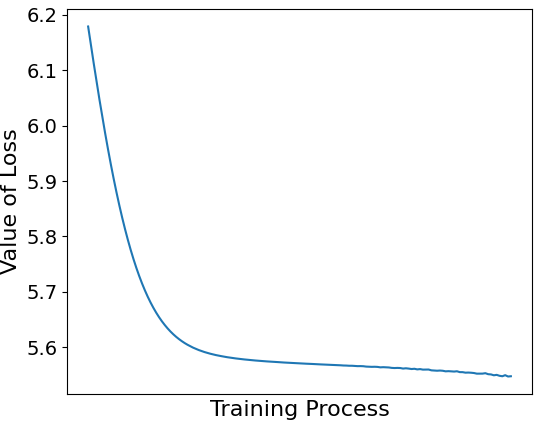}}
\centerline{(b) Encoder loss $\mathcal{L}_{\mathcal{F}}$ on BAT}
\vspace{3pt}
% \centerline{\includegraphics[width=0.9\textwidth]{BAT_Q_loss.png}}
% \centerline{(d) RL loss $\mathcal{L}_{\mathcal{Q}}$ on BAT}
\end{minipage}
\caption{Loss converge analysis on two datasets.}
\label{loss_function}
\end{figure}

\section{CONCLUSION}
In this paper, we find that the promising performance of the existing deep graph clustering methods heavily relies on the predefined cluster number $K$, which is not always known in practice. To solve this open problem, we propose a new deep non-parametric graph clustering method termed Reinforcement Graph Clustering (RGC) by learning the cluster number with reinforcement learning. In RGC, the states are built with both node and cluster embeddings to capture the local and global information in the graph. Subsequently, the designed quality network can evaluate the quality of cluster number (action) at the state. Moreover, a clustering-orient reward mechanism is proposed to improve the cohesion of the same clusters while separating the different clusters. In this manner, after optimization, the quality network can automatically determine the cluster number and achieve promising clustering performance. We hope this paper can motivate the researchers to design deeper non-parametric graph clustering methods. In the future, how to determine the cluster number in deep graph clustering methods based on the sample density will be an exciting topic.

% \appendix

\begin{acks}
This work was supported by the National Key R$\&$D Program of China 2020AAA0107100 and the National Natural Science Foundation of China (project no. 62325604, 62276271). Besides, this work was also supported by the National Key R\&D Program of China (Project 2022ZD0115100), the National Natural Science Foundation of China (Project U21A20427), the Research Center for Industries of the Future (Project WU2022C043), and the Competitive Research Fund (Project WU2022A009) from the Westlake Center for Synthetic Biology and Integrated Bioengineering.

\end{acks}

\newpage

\bibliographystyle{ACM-Reference-Format}
\bibliography{cite}
\end{document}

\title{Appendix of Paper "Reinforcement Graph Clustering with \\ Unknown Cluster Number"}

\renewcommand{\shortauthors}{Submission Id: 2024}

\maketitle

\appendix
\section*{Appendix}
Due to the limited pages of the main body, we introduce some details of our proposed Reinforcement Graph Clustering (RGC) and conduct additional experiments in the following sections.  

\section{Details of the Proposed Method}
In this section, we detail the architecture of encoder $\mathcal{F}$, loss function of $\mathcal{F}$, the architecture of quality network $\mathcal{Q}$, and the calculation process of complexity analysis.
\subsection{Architecture of encoder $\mathcal{F}$}
Given the node attributes $\textbf{X}$ and the original adjacency matrix $\textbf{A}$, we firstly perform Laplacian filtering for $\textbf{X}$ as follow. 

\begin{equation}
\widetilde{\textbf{X}} = (\prod_{i=1}^{t}(\textbf{I} - \widetilde{\textbf{L}}))\textbf{X} = (\textbf{I} - \widetilde{\textbf{L}})^t\textbf{X},
\label{SMOOTH}
\end{equation}
where $\widetilde{\textbf{L}}$ denotes the symmetric normalized graph Laplacian matrix and $t$ denotes the filtering times. Subsequently, $\widetilde{\textbf{X}}$ is embedded by $\text{Lin}_1$ and $\text{Lin}_2$ as follows.

\begin{equation} 
\begin{aligned}
\textbf{Z}^{v_1} &= \text{Lin}_1(\widetilde{\textbf{X}}); \textbf{Z}^{v_1}_i = \frac{\textbf{Z}^{v_1}_i}{||\textbf{Z}^{v_1}_i||_2}, i=1,2,...,N; \\
\textbf{Z}^{v_2} &= \text{Lin}_2(\widetilde{\textbf{X}});  \textbf{Z}^{v_2}_j = \frac{\textbf{Z}^{v_2}_j}{||\textbf{Z}^{v_2}_j||_2}, j=1,2,...,N.
\end{aligned}
\label{Lin}
\end{equation}
Here, $\textbf{Z}^{v_1}$ and $\textbf{Z}^{v_2}$ denote two-view node embeddings. For $\text{Lin}_1$ and $\text{Lin}_2$, they are both simple MLPs with same architecture but un-shared parameters, endow different semantics to two views. Then the output node embeddings can be calculated by linear combination of two views as follow. 
\begin{equation}
\textbf{Z} = (\textbf{Z}^{v_1} + \textbf{Z}^{v_2}) / 2.
\label{fusion}
\end{equation}

\subsection{Loss function of encoder $\mathcal{F}$}
The loss function of training encoder $\mathcal{F}$ includes two parts, i.e., contrastive loss $\mathcal{L}_{con}$ and clustering guidance loss $\mathcal{L}_{clu}$. To be specific, $\mathcal{L}_{con}$ is formulated as follows.

\begin{equation} 
\mathcal{L}_{con}=\frac{1}{2N}\sum_{j=1}^2\sum_{i=1}^N \mathcal{L}_i^{v_j},
\label{con_loss}
\end{equation} 

\begin{equation} 
\mathcal{L}_i^{v_j} = -log\frac{e^{\theta(i^{v_j}, i^{v_l})}}{e^{\theta(i^{v_j}, i^{v_l})}+\sum\limits_{k\neq i}({e^{\theta (i^{v_j},k^{v_j})}}+e^{\theta (i^{v_j}, k^{v_l})})}.
\label{infoNCE}
\end{equation}
where $j\neq l$. $\mathcal{L}_i^{v_j}$ denotes the loss for $i$-th node in $j$-th view. Besides, $\theta(\cdot)$ denotes the cosine similarity between the paired samples in the latent space. Namely, $\theta(i^{v_j},k^{v_l})=\textbf{Z}_i^{v_j}\textbf{Z}_k^{v_l}$. By minimizing $\mathcal{L}_{con}$, we pull together the same samples in different views while pushing away other samples.

In addition, the clustering algorithm is performed on the node embeddings $\textbf{Z}$ and obtain the cluster center $\textbf{C} \in \mathds{R}^{\hat{K}\times d}$ and cluster assignment matrix $\textbf{P}\in \mathds{R}^{N \times \hat{K}}$ as follow.

\begin{equation} 
\textbf{C}, \textbf{P} = \mathcal{P}(\textbf{Z}, \hat{K}),
\label{clustering}
\end{equation}
where the cluster center $\textbf{C}$ is calculated by averaging the node embeddings within each clusters. The clustering guidance loss $\mathcal{L}_{clu}$ is then formulated as follows.

\begin{equation} 
\textbf{G}_{ij} = \frac{(1+\|\textbf{Z}_i-\textbf{C}_j\|^2)^{-1}}{\sum_{j'}(1+\|\textbf{Z}_i-\textbf{C}_{j'}\|^2)^{-1}},
\label{G}
\end{equation} 

\begin{equation} 
\textbf{H}_{ij} = \frac{\textbf{G}_{ij}^2/\sum_i\textbf{G}_{ij}}{(\sum_{j'}\textbf{G}_{ij'}^2)/(\sum_i\textbf{G}_{ij})},
\label{H}
\end{equation}

\begin{equation} 
\mathcal{L}_{clu}=KL(\textbf{G}\|\textbf{H})=\sum_i \sum_j \textbf{G}_{ij}log\frac{\textbf{G}_{ij}}{\textbf{H}_{ij}}.
\label{clu_loss}
\end{equation} 
Here, $\textbf{G}$ and $\textbf{H}$ denote the clustering distribution and the sharpened clustering distribution. By minimizing $\mathcal{L}_{clu}$, we align the clustering distribution with the sharpened ones, thus improving the cluster cohesion. 

In summary, the total loss of encoder $\mathcal{L}_{\mathcal{F}}$ is formulated as follow. 
\begin{equation} 
\mathcal{L}_{\mathcal{F}}= \mathcal{L}_{con}+\alpha\mathcal{L}_{clu},
\label{f_loss}
\end{equation} 
where $\alpha$ denotes the trade-off hyper-parameter.

\subsection{Architecture of quality network $\mathcal{Q}$}
In this section, we detail the network architecture of quality network $\mathcal{Q}$. As mentioned in the main body, given the state $\textbf{S}_t$, $\mathcal{Q}$ will output the quality score vector $\textbf{q}_t$ as follow.
\begin{equation} 
\textbf{q}_t = \mathcal{Q}(\textbf{S}_t) = \mathcal{Q}(\{\textbf{Z}_t, \textbf{C}_t\}).
\label{quality}
\end{equation}
Concretely, we embed the states into the latent space as follow.
\begin{equation} 
\begin{aligned}
\textbf{Z'}_t &= \sigma(\text{Norm}(\text{Lin}_\textbf{Z}(\textbf{Z}_t))), \\
\textbf{C'}_t &= \sigma(\text{Norm}(\text{Lin}_\textbf{C}(\textbf{C}_t))),
\end{aligned}
\label{q_encoder}
\end{equation}
where, $\text{Lin}_{\textbf{Z}}$ and $\text{Lin}_{\textbf{C}}$ denote the MLPs for nodes and clusters, respectively. Besides, $\text{Norm}$ and $\sigma$ denote normalization and activate function, respectively. Then the node and cluster representations are concatenated together as follow.
\begin{equation} 
\textbf{O}_t = \text{Concat}([\textbf{Z'}_t, \textbf{C'}_t]).
\label{con}
\end{equation}
Eventually, one output layer $\text{Lin}_{\text{out}}$ and the softmax function are adopted to calculate the quality score vector as follow.
\begin{equation} 
\textbf{q}_t = \text{Softmax}(\text{Lin}_{\text{out}}(\textbf{O}_t)).
\label{con}
\end{equation}
By this settings, we firstly embed the collected node and cluster states into the deep latent space and then fusion them to learn the quality of different cluster numbers (actions) for each states. Moreover, after optimization, our RGC can automatically determine the cluster number for the clustering task and achieve promising performance.

\subsection{Calculation process of complexity analysis}
In the main body of this paper, we analysis the time and space complexity of calculating the loss functions in our proposed RGC. Concretely, assume that the max cluster number, the experience buffer size, and the encoding time for one state is $N_K$, $N_{\mathcal{B}}$, and $\text{T}_\mathcal{F}$, respectively. Besides, the state memory cost, the node number, and the latent feature dimensions denotes $M_{\textbf{S}}$, $N$, and $d$, respectively. Thus, the time complexity of calculating $\mathcal{L}_{\mathcal{Q}}$, $\mathcal{L}_{con}$, $\mathcal{L}_{clu}$ is $\mathcal{O}(N_{\mathcal{B}}\text{T}_\mathcal{F}+N_{\mathcal{B}}N_K/2+N_{\mathcal{B}}/2)$, $\mathcal{O}(N^2d)$, and $\mathcal{O}(N\hat{K})$, respectively. In addition, the space complexity of $\mathcal{L}_{\mathcal{Q}}$, $\mathcal{L}_{con}$, $\mathcal{L}_{clu}$ is $\mathcal{O}(N_{\mathcal{B}}M_{\textbf{S}})$, $\mathcal{O}(N^2)$, and $\mathcal{O}(N\hat{K})$, respectively.

In this section, we detail the calculation process of the analysis. Firstly, for the Reinforcement learning loss $\mathcal{Q}$, it is formulated as follow.
\begin{equation} 
\mathcal{L}_{\mathcal{Q}} = \frac{1}{t_e-t_s} \sum_{t=t_s}^{t_e-1} (R_t+\gamma \cdot max \mathcal{Q}(\textbf{S}_{t+1}) - \mathcal{Q}(\textbf{S}_{t})[\hat{K}_t])^2.
\label{Q_loss}
\end{equation}
To calculate $\mathcal{Q}$, we need to encode the states for twice, namely $\mathcal{Q}(\textbf{S}_{t+1})$ and $\mathcal{Q}(\textbf{S}_{t})$, costing $2\text{T}_\mathcal{F}$ time. Besides, to find the best action at state $\textbf{S}_{t+1}$, it costs $N_K$ time. Moreover, for other operations like, addition, subtraction and multiplication, they take the constant time cost $\mathcal{O}(1)$. In summary, for $N_{\mathcal{B}}$ states in one buffer, the total time costs are formulated as follow.
\begin{equation}
\begin{aligned}
&&\mathcal{O}(N_{\mathcal{B}}(2\text{T}_{\mathcal{F}}+N_K+1))  = \\ &&\mathcal{O}(2N_{\mathcal{B}}\text{T}_{\mathcal{F}}+N_{\mathcal{B}}N_K+N_{\mathcal{B}})	  \approx \\ && \mathcal{O}(N_{\mathcal{B}}\text{T}_\mathcal{F}+N_{\mathcal{B}}N_K/2+N_{\mathcal{B}}/2).
\end{aligned}
\label{analysis_Q}
\end{equation}
During the process, it needs to keep the memory of the buffer, thus costing $N_B M_{\textbf{S}}$. For the contrastive loss $\mathcal{L}_{con}$, it is formulated in Eq. \eqref{con_loss} and Eq. \eqref{infoNCE}. There, the similarity calculation for each sample pair will cost $N^2d$ time. Other operations like sum and division cost constant time. Thus, the total time cost of $\mathcal{L}_{con}$ is formulated as $\mathcal{O}(N^2d)$. During this process, the memory cost is $N^2$ since we need to keep the similarities between each sample pair. Furthermore, for the clustering loss $\mathcal{L}_{clu}$, it is formulated in Eq. \eqref{G}, Eq. \eqref{H} and Eq. \eqref{clu_loss}. There, it needs to calculate the distance between nodes with the cluster centers, costing $N\hat{K}$ time. $\hat{K}$ is the learned cluster number. Besides, other operations like addition, sum, and division cost constant time. Thus, the time cost of $\mathcal{L}_{clu}$ is $\mathcal{O}(N \hat{K})$. In this process, we need to keep the distance between nodes and cluster centers in both $\textbf{G}$ and $\textbf{H}$, thus costing $\mathcal{O}(2N\hat{K}) \approx \mathcal{O}(N\hat{K})$ memory.

% The detailed calculation process can be found in Appendix A.4.

% \bibliography{cite}

\section{Additional Experimental Result}
In this section, we show the detailed statistics of datasets and the additional experimental results.
\subsection{Statistics}
The statistics of benchmark datasets are summarized in Table \ref{DATASET_INFO}.
\begin{table}[h]
\centering
\small
\caption{Statistics of six benchmark datasets.}
\scalebox{0.9}{
\begin{tabular}{c|cclclclclcl}
\hline
                                     & \textbf{Dataset} & \multicolumn{2}{c}{\textbf{CORA}} & \multicolumn{2}{c}{\textbf{CITESEER}} & \multicolumn{2}{c}{\textbf{AMAP}} & \multicolumn{2}{c}{\textbf{BAT}} & \multicolumn{2}{c}{\textbf{EAT}} \\ \hline
\multirow{5}{*}{\textbf{Statistics}} & Type             & \multicolumn{2}{c}{Graph}         & \multicolumn{2}{c}{Graph}         & \multicolumn{2}{c}{Graph}         & \multicolumn{2}{c}{Graph}        & \multicolumn{2}{c}{Graph}        \\
                                     & \# Samples       & \multicolumn{2}{c}{2708}          & \multicolumn{2}{c}{3327}          & \multicolumn{2}{c}{7650}          & \multicolumn{2}{c}{131}          & \multicolumn{2}{c}{399}          \\
                                     & \# Dimensions    & \multicolumn{2}{c}{1433}          & \multicolumn{2}{c}{3703}          & \multicolumn{2}{c}{745}           & \multicolumn{2}{c}{81}           & \multicolumn{2}{c}{203}          \\
                                     & \# Edges         & \multicolumn{2}{c}{5429}          & \multicolumn{2}{c}{4732}          & \multicolumn{2}{c}{119081}        & \multicolumn{2}{c}{1038}         & \multicolumn{2}{c}{5994}         \\
                                     & \# Classes       & \multicolumn{2}{c}{7}             & \multicolumn{2}{c}{6}             & \multicolumn{2}{c}{8}             & \multicolumn{2}{c}{4}            & \multicolumn{2}{c}{4}            \\ \hline
\end{tabular}}
\label{DATASET_INFO}
\end{table}

\subsection{Hyper-parameter analysis}

\begin{figure}[h]
\centering
\small
\begin{minipage}{0.48\linewidth}
\centerline{\includegraphics[width=0.9\textwidth]{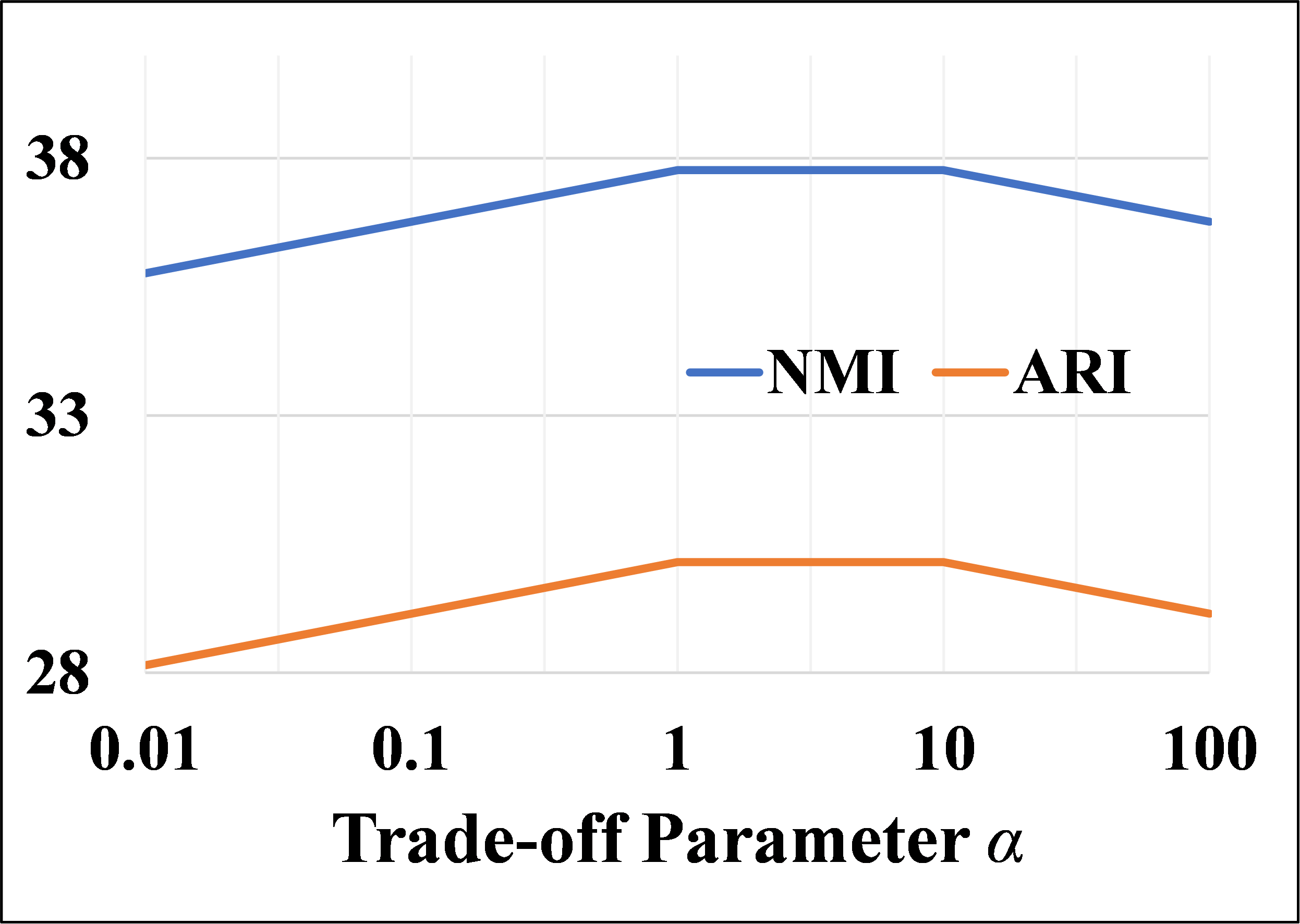}}
\centerline{(a) $\alpha$ on EAT}
\end{minipage}
\begin{minipage}{0.48\linewidth}
\centerline{\includegraphics[width=0.9\textwidth]{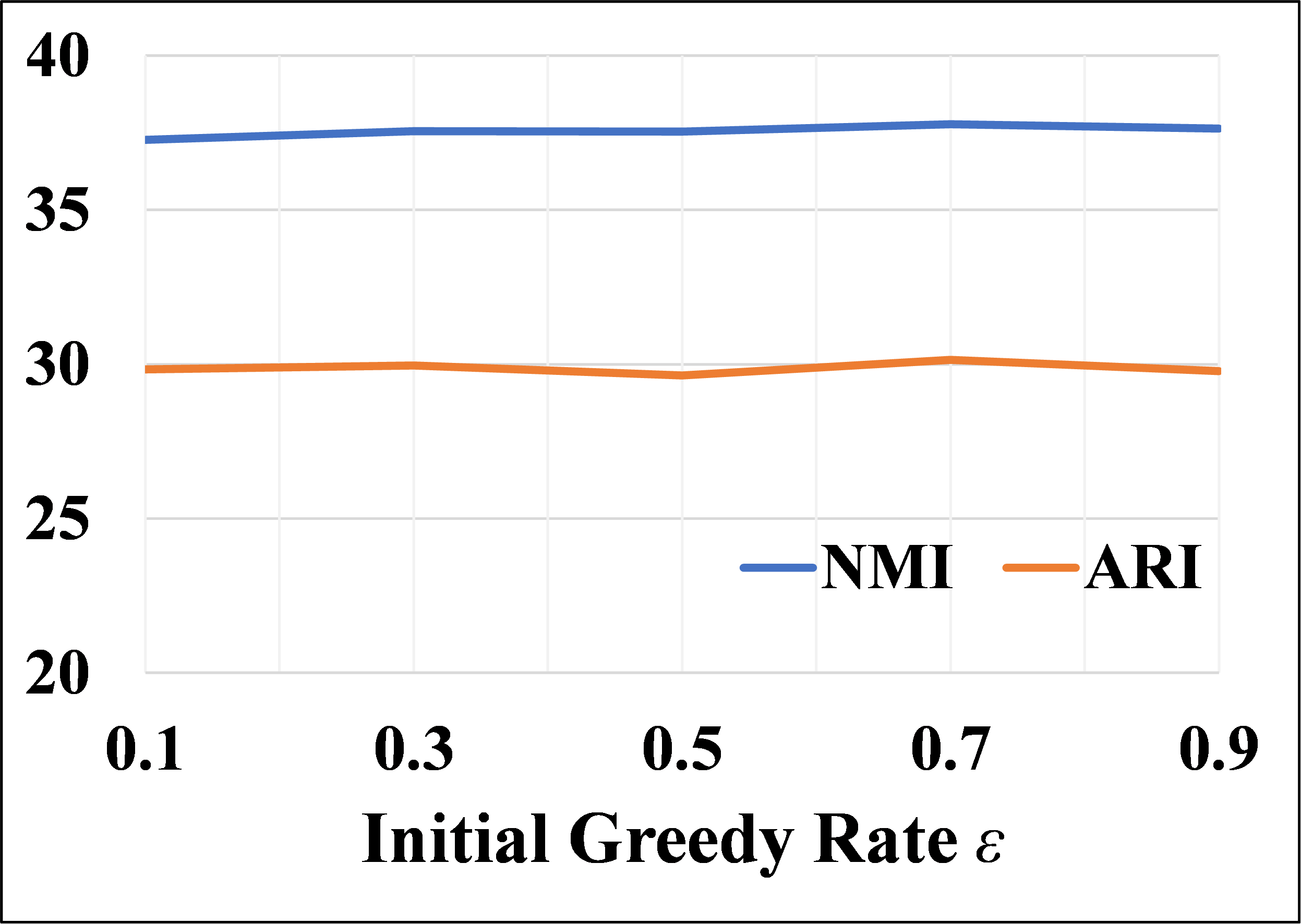}}
\centerline{(b) $\epsilon$ on EAT}
\end{minipage}
\caption{Hyper-parameter analysis of $\alpha$ and $\epsilon$ on EAT dataset.}
\label{eat_hyper}
\end{figure}

\begin{figure}[h]
\centering
\small
\begin{minipage}{0.48\linewidth}
\centerline{\includegraphics[width=0.9\textwidth]{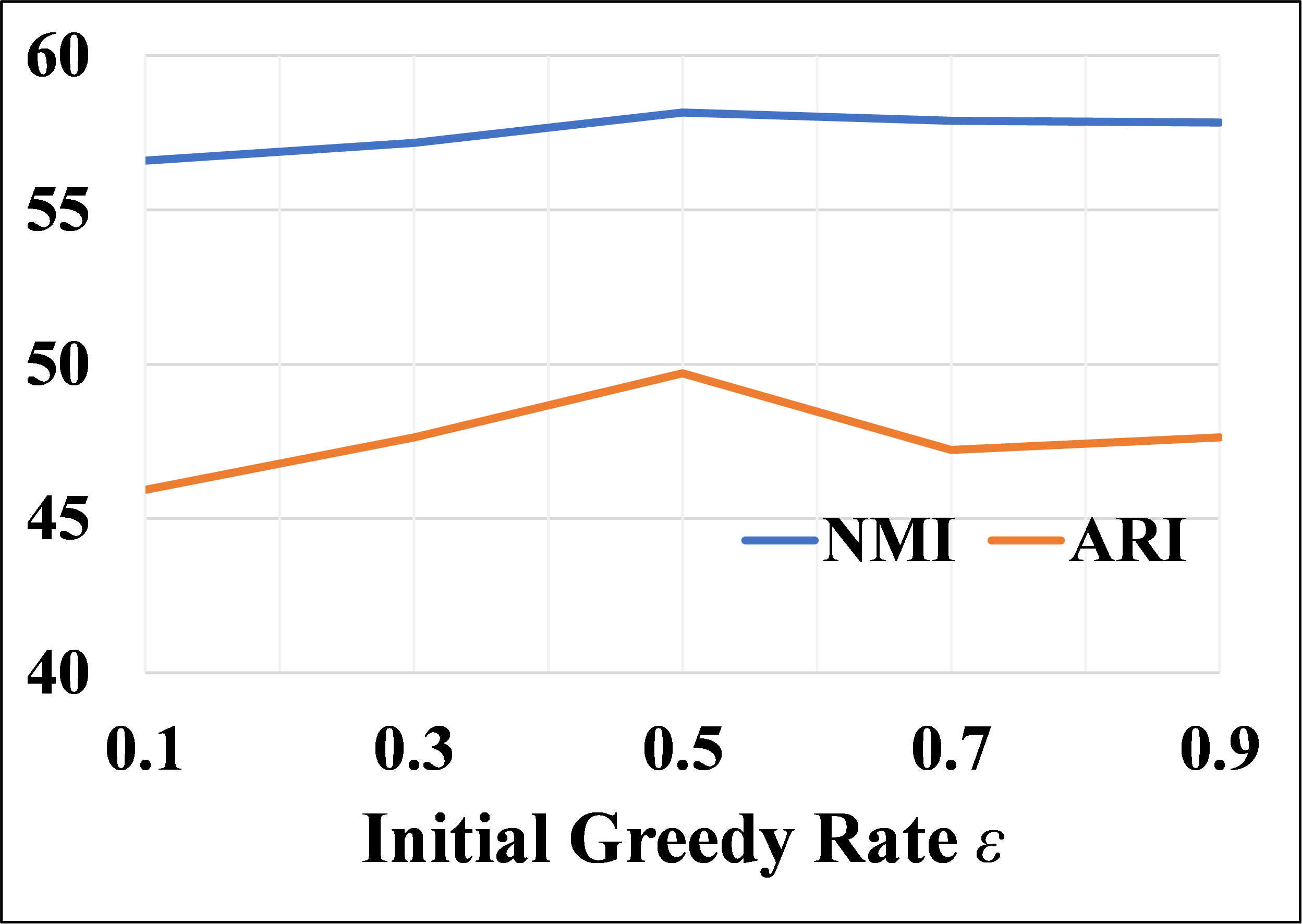}}
\centerline{(a) CORA}
\centerline{\includegraphics[width=0.9\textwidth]{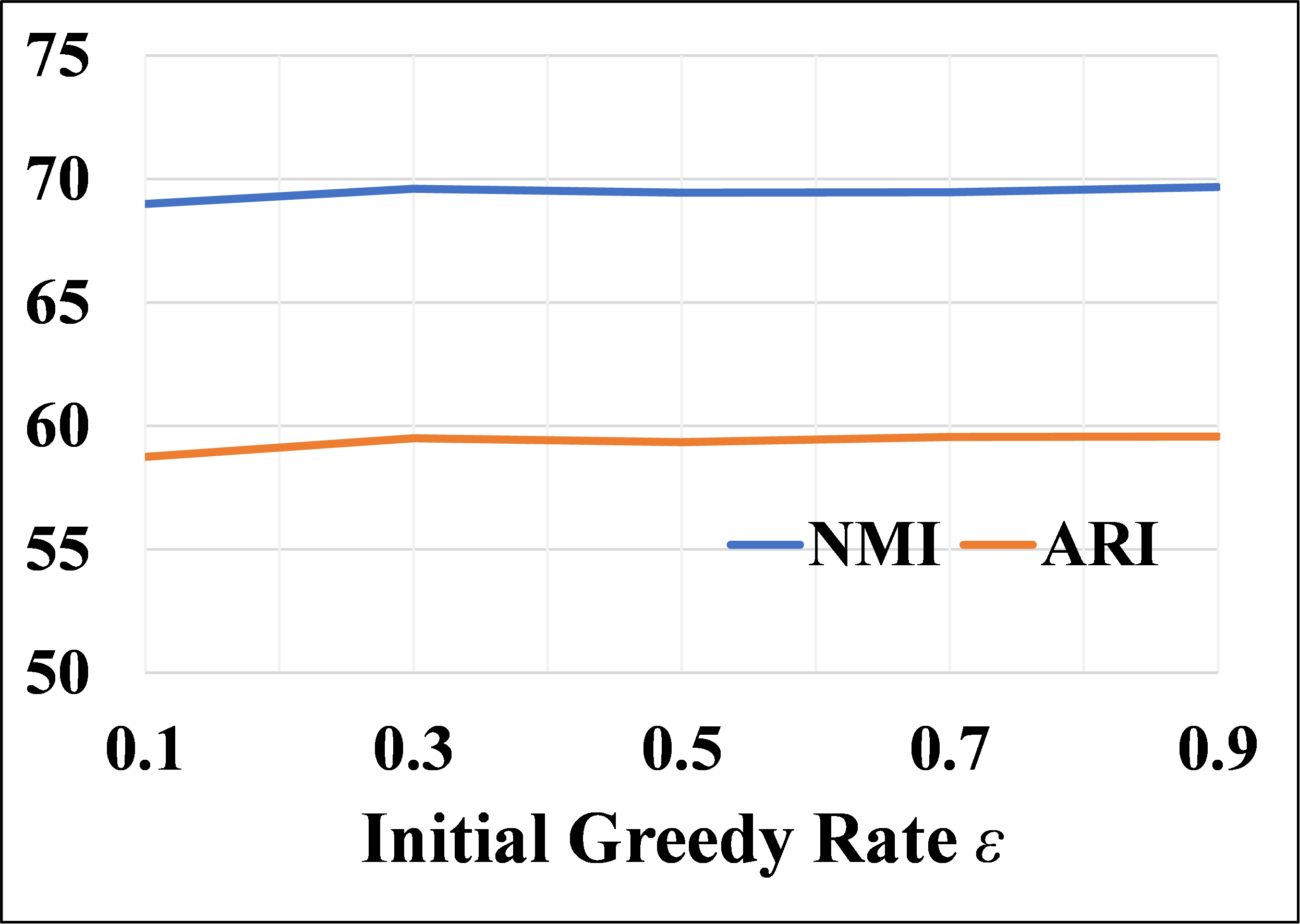}}
\centerline{(c) AMAP}
\end{minipage}
\begin{minipage}{0.48\linewidth}
\centerline{\includegraphics[width=0.9\textwidth]{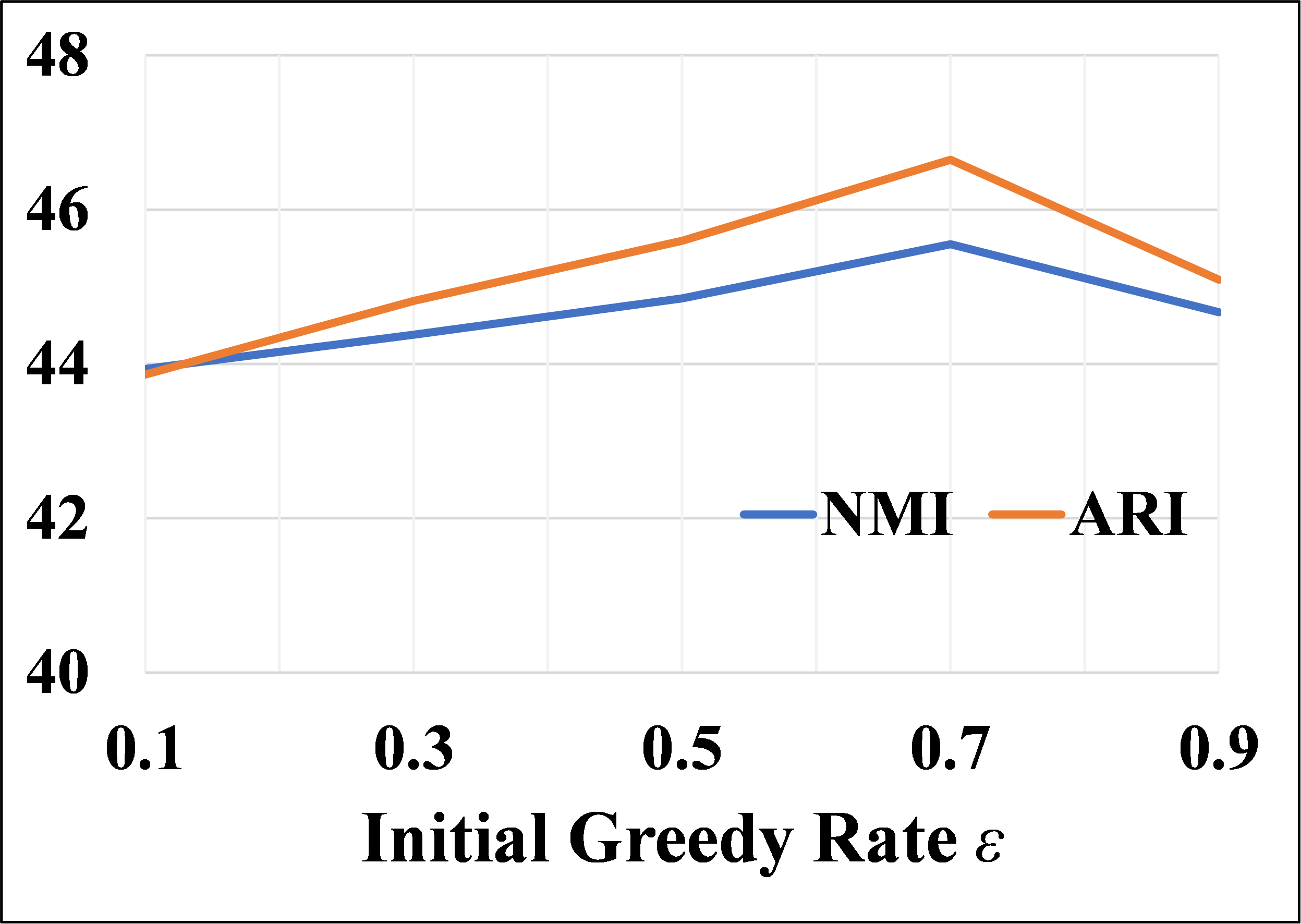}}
\centerline{(b) CITESEER}
\centerline{\includegraphics[width=0.9\textwidth]{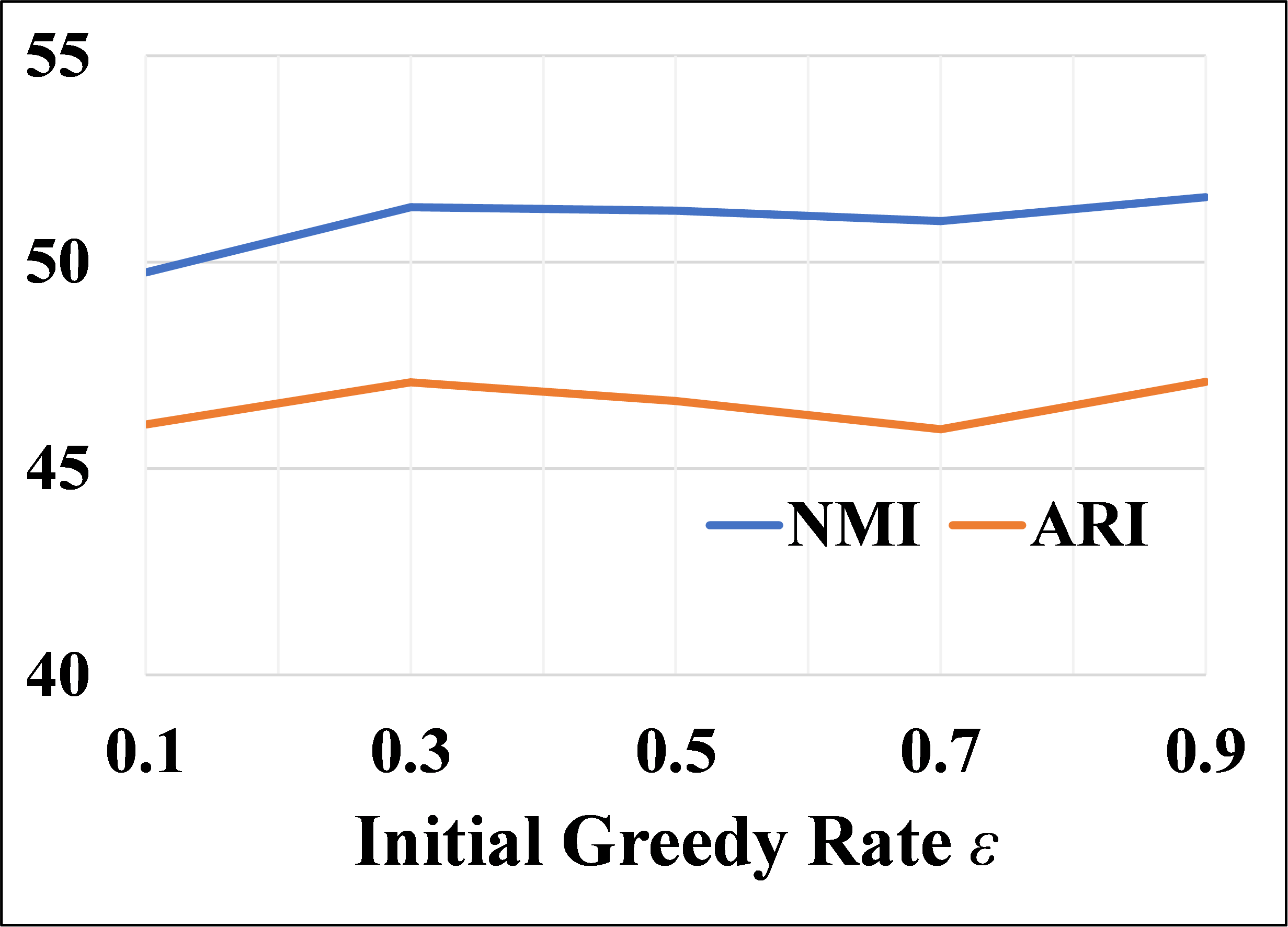}}
\centerline{(d) BAT}
\end{minipage}
% \begin{minipage}{0.48\linewidth}
% \centerline{\includegraphics[width=0.9\textwidth]{epsilon_eat.png}}
% \centerline{(e) EAT}
% \end{minipage}
\caption{Sensitivity analysis of the initial greedy rate hyper-parameter $\epsilon$ on four datasets.}
\label{epsilon}
\end{figure}

\begin{figure}[h]
\centering
\small
\begin{minipage}{0.48\linewidth}
\centerline{\includegraphics[width=0.9\textwidth]{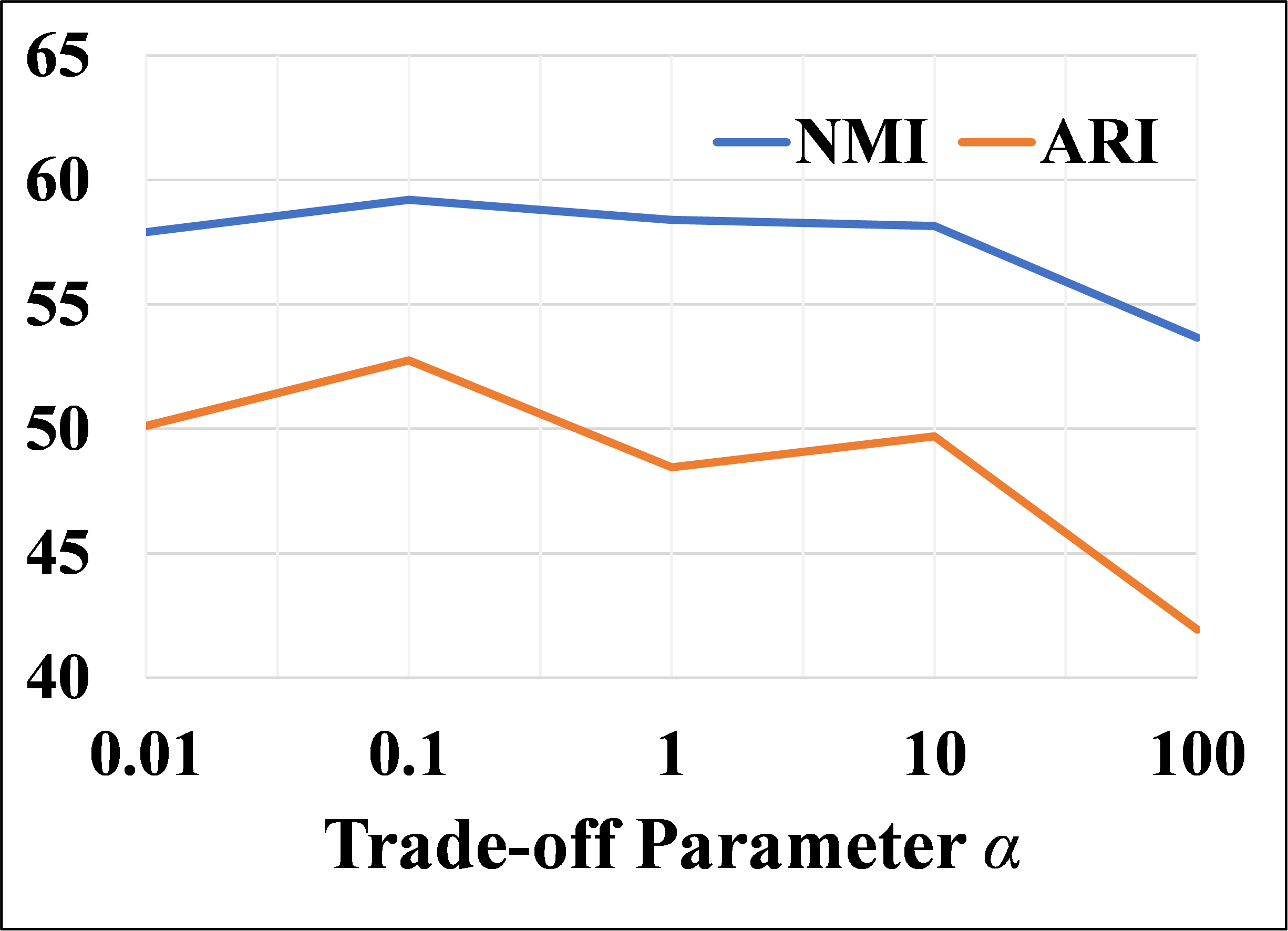}}
\centerline{(a) CORA}
\centerline{\includegraphics[width=0.9\textwidth]{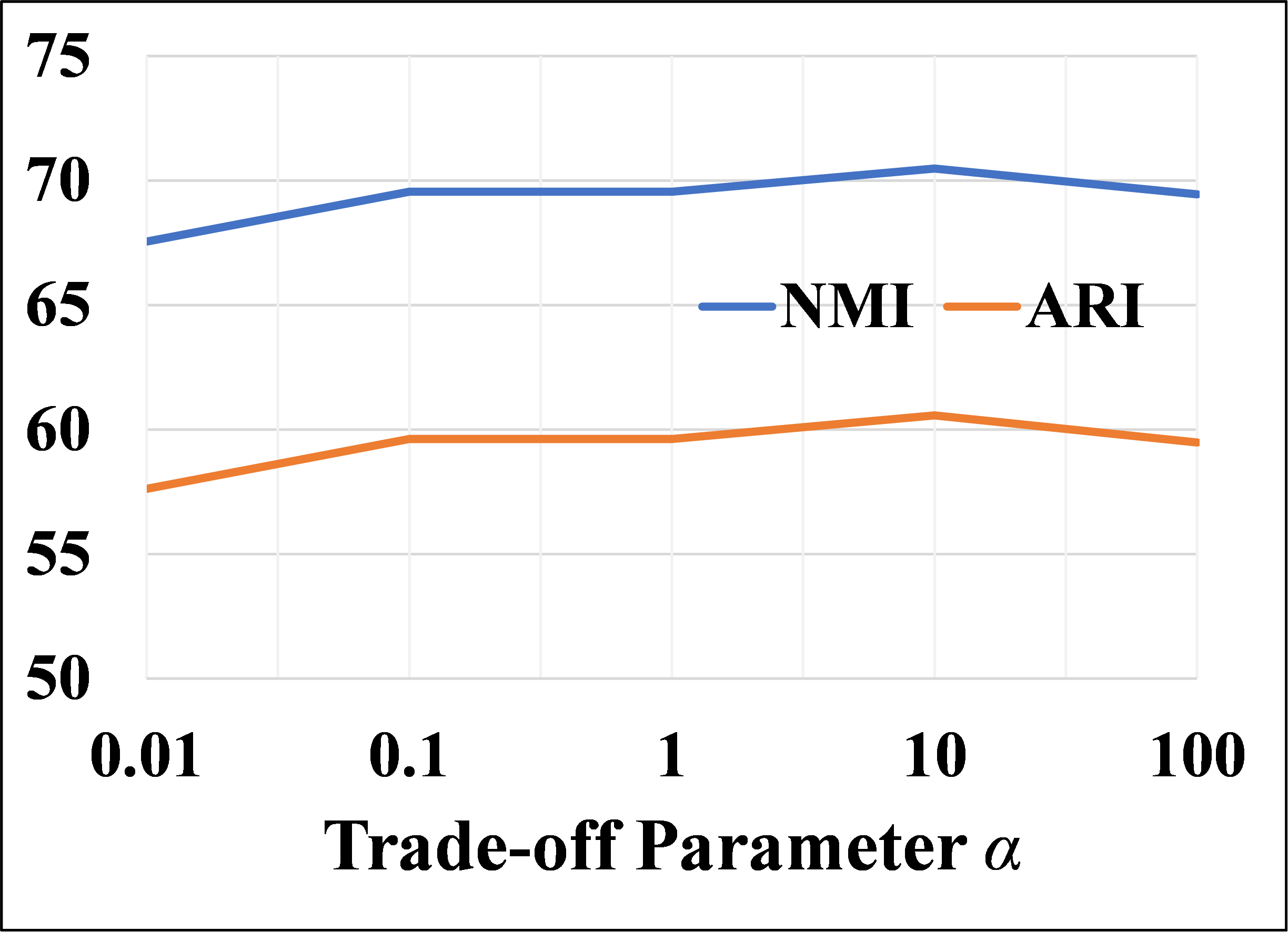}}
\centerline{(c) AMAP}
\end{minipage}
\begin{minipage}{0.48\linewidth}
\centerline{\includegraphics[width=0.9\textwidth]{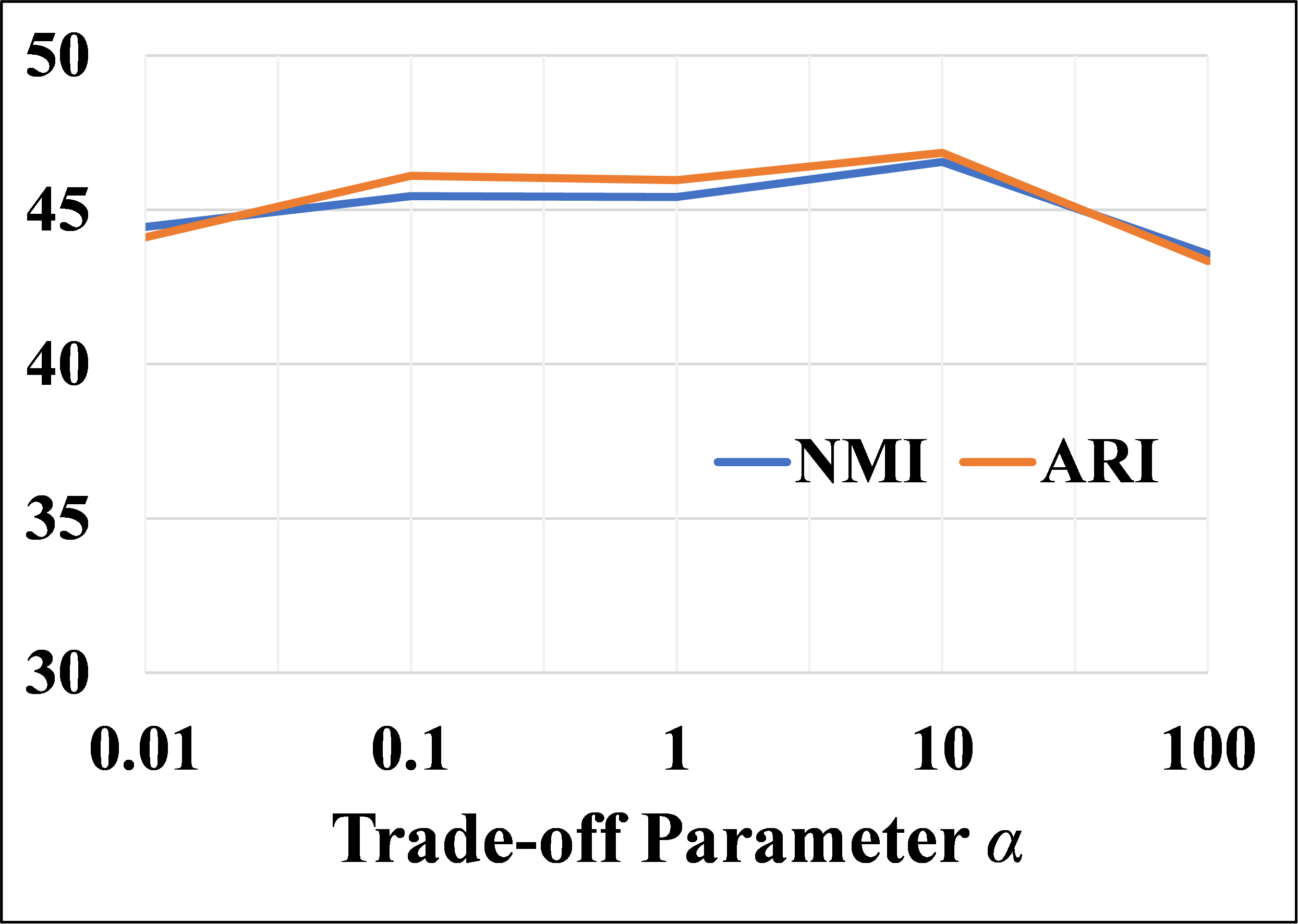}}
\centerline{(b) CITESEER}
\centerline{\includegraphics[width=0.9\textwidth]{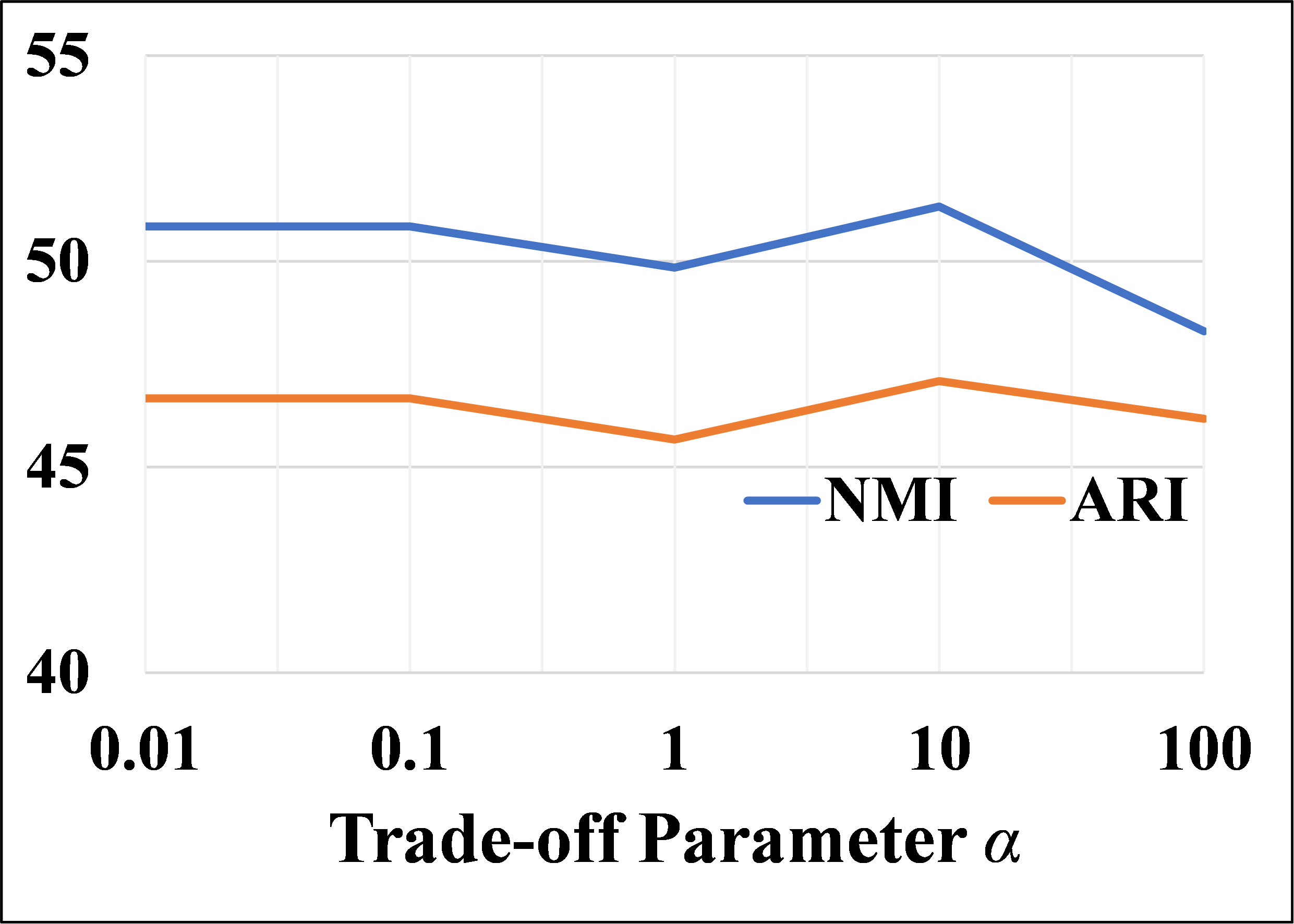}}
\centerline{(d) BAT}
\end{minipage}
% \begin{minipage}{0.48\linewidth}
% \centerline{\includegraphics[width=0.9\textwidth]{alpha_eat.png}}
% \centerline{(e) EAT}
% \end{minipage}
\caption{Hyper-parameter analysis of the trade-off parameter $\alpha$ on four datasets.}
\label{alpha}
\end{figure}

We analyze the sensitivity of hyper-parameters, including $\epsilon$ and $\alpha$ in our proposed RGC. Specifically, $\epsilon$ is the initial greedy rate and $\alpha$ denotes the trade-off parameter in $\mathcal{L}_{\mathcal{F}}$. As shown in Figure \ref{epsilon}, we conduct the sensitivity analysis experiments of hyper-parameter $\epsilon$ and have two conclusions as follows. 1) Our proposed RGC is not sensitive to the initial greedy rate $\epsilon$. It indicates that our method is not sensitive to initialization and will be optimized to achieve good clustering performance. 2) RGC can achieve promising performance when $\epsilon \in [0.3, 0.7]$. Thus, in our method, we search $\epsilon$ in $\{0.3,0.5,0.7\}$. Besides, as shown in Figure \ref{alpha}, we analyze the trade-off parameter $\alpha$ on four datasets and have two findings as follows. 1) Our proposed method is not sensitive to the trade-off parameter $\alpha$ when $\alpha \in [0.1, 10]$. 2) RGC will achieve promising performance on CITESEER, AMAP, and BAT datasets when $\alpha=10$. Thus, in our proposed method, $\alpha$ is set to 10. The additional experimental results of the sensitivity analysis of hyper-parameter $\epsilon$ and $\alpha$ is shown in Figure \ref{eat_hyper}. Similar conclusions with that in the main body of the paper can be obtained on EAT dataset. 

\begin{figure}[h]
\centering
\small
\begin{minipage}{0.49\linewidth}
\centerline{\includegraphics[width=0.9\textwidth]{CORA_E_loss.png}}
\centerline{(a) Encoder loss $\mathcal{L}_{\mathcal{F}}$ on CORA}
\vspace{3pt}
\centerline{\includegraphics[width=0.9\textwidth]{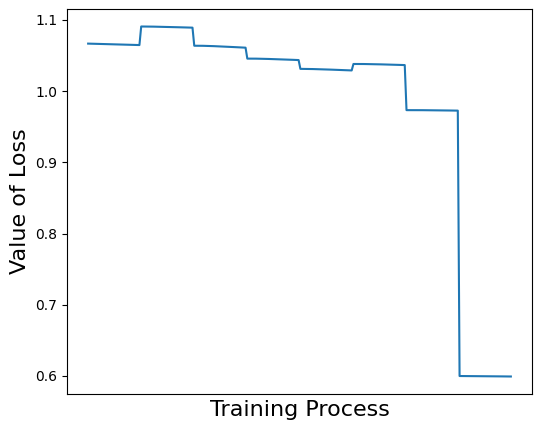}}
\centerline{(c) RL loss $\mathcal{L}_{\mathcal{Q}}$ on CORA}
\end{minipage}
\begin{minipage}{0.49\linewidth}
\centerline{\includegraphics[width=0.9\textwidth]{BAT_E_loss.png}}
\centerline{(b) Encoder loss $\mathcal{L}_{\mathcal{F}}$ on BAT}
\vspace{3pt}
\centerline{\includegraphics[width=0.9\textwidth]{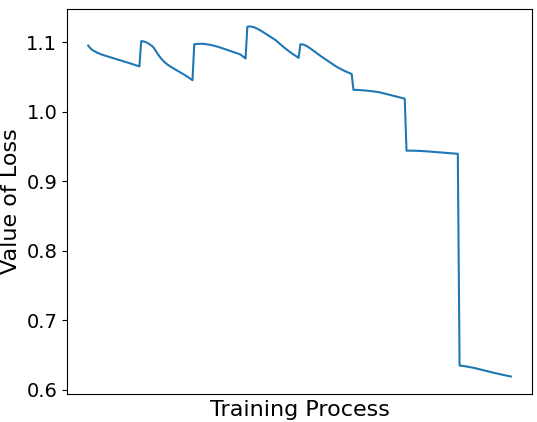}}
\centerline{(d) RL loss $\mathcal{L}_{\mathcal{Q}}$ on BAT}
\end{minipage}
\caption{Loss converge analysis on two datasets.}
\label{loss_function}
\end{figure}

The loss functions of our proposed RGC include two parts, i.e., the encoder loss $\mathcal{L}_{\mathcal{F}}$ for training encoder $\mathcal{F}$ and the reinforcement loss $\mathcal{L}_{\mathcal{Q}}$ for training the quality network $\mathcal{Q}$. In this section, we plot the curve of loss functions on two datasets, CORA and BAT, as shown in Figure \ref{loss_function}. From these visualization results, the conclusion is that both $\mathcal{L}_{\mathcal{F}}$ and $\mathcal{L}_{\mathcal{Q}}$ can decrease during training and eventually converge.

% \bibliographystyle{ACM-Reference-Format}
% \bibliography{cite}